\documentclass[final,5p,times,twocolumn]{elsarticle}

\usepackage{lineno,hyperref}
\modulolinenumbers[5]

\journal{arXiv}

\usepackage{times}
\usepackage{epsfig}
\usepackage{graphicx}
\usepackage{amsmath}
\usepackage{amssymb}

\usepackage{tabularx}
\usepackage{multirow}
\usepackage{hhline}
\usepackage{booktabs}

\usepackage{epstopdf}
\usepackage{subfig}
\usepackage{color}

\makeatletter
\newcommand*{\rom}[1]{\expandafter\@slowromancap\romannumeral #1@}
\makeatother
\newcommand{\etal}{\textit{et al}.}
\newcommand{\eg}{\textit{e}.\textit{g}.}
\newcommand{\etc}{\textit{etc}.}
\newcommand{\ie}{\textit{i}.\textit{e}.}

\begin{document}

\let\today\relax

\begin{frontmatter}

\title{Generic Instance Search and Re-identification from One Example via Attributes and Categories}


\author[mysecondaryaddress]{Ran Tao\corref{mycorrespondingauthor}}
\cortext[mycorrespondingauthor]{Corresponding author}
\ead{R.Tao@uva.nl}

\author[mysecondaryaddress]{Arnold W.M. Smeulders}

\author[mythirdaddress]{Shih-Fu Chang}

\address[mysecondaryaddress]{ISIS, University of Amsterdam, Science Park 904, Amsterdam, the Netherlands}
\address[mythirdaddress]{Department of Electrical Engineering, Columbia University, 500 W. 120th St., Mudd 1310, New York, USA}

\begin{abstract}
	This paper aims for generic instance search from one example where the instance can be an arbitrary 3D object like shoes, not just near-planar and one-sided instances like buildings and logos. First, we evaluate state-of-the-art instance search methods on this problem. We observe that what works for buildings loses its generality on shoes. Second, we propose to use automatically learned category-specific attributes to address the large appearance variations present in generic instance search. Searching among instances from the same category as the query, the category-specific attributes outperform existing approaches by a large margin on shoes and cars and perform on par with the state-of-the-art on buildings. Third, we treat person re-identification as a special case of generic instance search. On the popular VIPeR dataset, we reach state-of-the-art performance with the same method. Fourth, we extend our method to search objects without restriction to the specifically known category. We show that the combination of category-level information and the category-specific attributes is superior to the alternative method combining category-level information with low-level features such as Fisher vector.

	\textit{This technical report is an extended version of our previous conference paper ``Attributes and Categories for Generic Instance Search from One Example'' (CVPR 2015).}
\end{abstract}


\end{frontmatter}


\section{Introduction}
In instance search, the objective is to retrieve all images of a specific object given a few query examples of that object~\cite{arandjelovic2012multiple,jegou2008hamming,over2012trecvid,qin2011hello,zhu2013query}. We consider the challenging case of only 1 query image and admitting large differences in the imaging angle and other imaging conditions between the query image and the target images. A very hard case is a query specified in frontal view while the relevant images in the search set show a view from the back which has never been seen before. Humans solve the search task by employing two types of general knowledge. First, when the query instance is a certain class, say a female, answers should be restricted to be from the same class. And, queries in the frontal view showing one attribute, say brown hair, will limit answers to show the same attribute, even when the viewpoint is from the back. In this paper, we exploit these two types of knowledge to handle a wide variety of viewpoints, illumination and other conditions for instance search. 

In instance search, excellent results have been achieved by restricting the search to buildings~\cite{arandjelovic2012multiple,arandjelovic2012three,philbin2007object,gavves2012visual}. Searching buildings can be used in location recognition and 3D reconstruction. Another set of good results has been achieved in searching for logos~\cite{joly2009logo,revaud2012correlation,tao2014locality} for the estimation of brand exposure. And,~\cite{wang2011contextual} searches for book and magazine covers. 
All these cases of instance search show good results for near-planar, and one-sided objects which are recorded under a limited range of imaging angles. In this work, we aim for broader classes of query instances. We aim to perform generic instance search from 1 example. \textit{Generic} implies we consider arbitrary objects, and not just one-sided objects. And, \textit{generic} implies we aim to use one approach not specially designed for a certain kind of instances, such as RANSAC-based geometric verification for rigid and highly textured objects. In our case, instances can be buildings and logos, but also shoes, clothes and other objects. In this paper, we illustrate on a diverse set of instances, including shoe, car, building and person. 
In this work, we treat person re-identification~\cite{gong2014person} as a special case of generic instance search, and address the problem using the same method as for other kinds of instances.

The challenge in instance search is to represent the query image invariant to the (unknown) appearance variations of the query while maintaining a sufficiently rich representation to permit distinction from other, similar instances. To solve this, most existing approaches in instance search match the appearance of local spots~\cite{lowe2004distinctive,bay2008speeded} in the potential target to the query~\cite{jegou2008hamming,jegou2014triangulation,philbin2007object,tao2014locality,tolias2013aggregate}. The quality of match in these approaches between two images is the sum of similarities over all local descriptor pairs. The difference between the cited approaches lies in the way local descriptors are encoded and in the computation of the similarity. Good performance has been achieved by this paradigm on buildings, logos and scenes from a distance. However, when searching for an arbitrary object with a wider range of viewpoint variability, more sides, and possibly having self-occlusion and non-rigid deformation, these methods are likely to fail as local descriptor matching becomes unreliable in these cases~\cite{mikolajczyk2005performance}. 


In this paper we propose to use automatically learned attributes~\cite{farhadi2009describing,lampert2009learning} to address generic instance search. Attributes, as higher level abstractions of visual properties, have been shown advantageous in classification when training examples are insufficiently covering the variations in the original feature space~\cite{farhadi2009describing,yu2013designing,akata2013label}, surely present in the one-example challenging case. By employing attributes, we aim to be robust against intra-instance appearance variations. Further, we optimize the attributes such that they are meanwhile discriminative among different instances. Concretely, in this paper, we learn a set of category-specific non-semantic attributes that are optimized to recognize different instances of a certain category, \eg, shoes. With the learned attributes, an instance can be represented as a specific combination of the attributes, and instance search boils down to finding the most similar combinations of attributes.

In order to address the possible confusion of the query with instances from other categories, we further propose to supplement the learned category-specific attributes with category-level information. The category-level information are incorporated to reduce the search space by filtering instances of other categories. 
It is advantageous when there is only 1 query image, to use slightly more user provided information. In addition to the interactive specification of the object region in the query image, we require the specification of the category the query instance belongs to.

 

\section{Related work}
Most approaches in instance search rely on gathering matches of local image descriptors~\cite{sivic2003video,philbin2007object,jegou2008hamming,tolias2013aggregate,tao2014locality,jegou2014triangulation}, where the differences reside in the way the local descriptors are encoded and the matching score of two descriptors is evaluated. Bag-of-words (BoW)~\cite{sivic2003video,philbin2007object} encodes a local descriptor by the index of the nearest visual word. Hamming embedding~\cite{jegou2008hamming} improves upon BoW by adding an extra binary code to better describe the position of the local descriptor in space. The matching score of a pair of descriptors is 1 if they are encoded to the same word and the Hamming distance between binary signatures is smaller than a certain threshold. VLAD~\cite{jegou2010aggregating} and Fisher vector~\cite{perronnin2010large} improve over BoW by representing the local descriptor with an extra residual vector, obtained by subtracting the mean of the visual word or the Gaussian component respectively. In VLAD and Fisher vector, the score of two descriptors is the dot product of the residuals when they are encoded to the same word, and 0 otherwise.~\cite{tolias2013aggregate,tao2014locality} improve VLAD and Fisher vector by replacing the dot product by a thresholded polynomial similarity and an exponential similarity respectively to give disproportionally more credits to closer descriptor pairs.~\cite{jegou2014triangulation} encodes a local descriptor by only considering the directions to the visual word centers, not the magnitudes, outperforming Fisher vector on instance search. 
With these methods, good performance has been achieved on buildings, logos, and scenes from a distance. These instances can be conceived as near-planar and one-sided. For buildings, logos, and scenes from a distance the variation in the viewing angle is limited to a quadrant of 90 degrees at most out of the full 360 circle. For limited variations in viewpoint, matches of local descriptors can be reliably established between the query and a relevant example.
In this work, we consider generic instance search, where the instance can be an arbitrary object with a wider range of viewpoint variability and more sides. We evaluate existing methods for approximately one-sided instance search on this problem of generic instance search. 




Attributes~\cite{farhadi2009describing,ferrari2008learning,lampert2009learning} have received much attention recently. They are used to represent common visual properties of different objects. Attribute representation has been used for image classification~\cite{farhadi2009describing,yu2013designing,akata2013label}. Attributes have been shown to be advantageous when the training examples are insufficiently covering the appearance variations in the original feature space~\cite{farhadi2009describing,yu2013designing}. Inspired by this, we propose to use attribute representation to address generic instance search, where there is only 1 example available and there still exists a wide range of appearance variations. 


Attributes have been used for image retrieval~\cite{siddiquie2011image,kovashka2012whittlesearch,yu2012weak,yu2013designing,rastegari2012attribute}. In~\cite{siddiquie2011image,kovashka2012whittlesearch,yu2012weak}, the query is defined by textual attributes instead of images and the goal is to return images exhibiting query attributes. 
In the references, the query attributes need to be semantically meaningful such that the query can be specified by text. In this work, we address instance search given one query image, which is a different task as the correct answers have to exhibit the same instance (not just the same attributes), and we use automatically learned attributes which as a consequence may or may not be semantic.
~\cite{yu2013designing,rastegari2012attribute} consider non-semantic attributes for category retrieval, while this work addresses generic instance retrieval.

The use of category-level information to improve instance search has been explored in~\cite{zhang2013semantic,douze2011combining,gordoa2012leveraging}.~\cite{gordoa2012leveraging} uses category labels to learn a projection to map the original feature to a lower-dimensional space such that the lower-dimensional feature incorporates certain category-level information. 
In this work, instead of learning a feature mapping, we augment the original representation with additional features to capture the category-level information. 
In~\cite{douze2011combining}, Fisher vector representation is expanded with the concept classifier output vector of the 2659 concepts from Large Scale Concept Ontology for Multimedia (LSCOM)~\cite{naphade2006large}. In~\cite{zhang2013semantic}, a 1000-dimensional concept representation~\cite{LSVRC2010} is utilized to refine the inverted index on the basis of semantic consistency between images.
Both~\cite{zhang2013semantic} and~\cite{douze2011combining} combine category-level information with low-level representation. 
In this work, we consider the combination of category-level information with category-specific attributes rather than a low-level representation. We argue this is a more principled combination as the category-level information by definition makes category-level distinction and the category-specific attributes are optimized for within-category discrimination.


Person re-identification is a well-studied topic~\cite{gong2014person,bedagkar2014survey,vezzani2013people}, where the work mainly branches into two aspects, feature designing~\cite{gray2008viewpoint,ma2012local,yang2014salient} and metric learning~\cite{hirzer2012relaxed,chen2015similarity,paisitkriangkrai2015learning}. Among the vast amount of work in literature, most related to this paper are papers focusing on building a good representation~\cite{gray2008viewpoint,ma2012local,bazzani2013symmetry,zhao2013unsupervised,zhao2014learning,yang2014salient,liao2015person,ahmed2015improved,shi2015transferring}.~\cite{gray2008viewpoint} uses AdaBoost to select features from an ensemble of localized features.~\cite{ma2012local} encodes the local descriptors using Fisher vector.~\cite{bazzani2013symmetry} exploits the symmetry and asymmetry properties of human body to capture the cues on the human body only, pruning out background clutters.~\cite{zhao2013unsupervised} learns human saliency in an unsupervised manner to find reliable and discriminative patches.~\cite{zhao2014learning} proposes to learn mid-level patch filters that are viewpoint invariant and discriminative in differentiating identities.~\cite{yang2014salient} employs a salient color names based representation.~\cite{liao2015person} records the maximal local occurrence of a pattern to achieve invariance to viewpoint changes.~\cite{ahmed2015improved} simultaneously learns features and a similarity metric using deep learning.~\cite{shi2015transferring} proposes to learn semantic fashion-related attribute representation from auxiliary datasets and adapt the representation to target datasets. In this work, we propose to learn a non-semantic attribute representation without using auxiliary data to handle the large appearance variations caused by viewpoint differences, illumination variations, deformation and others. Furthermore, in this paper, inspired by~\cite{zheng2015query}, we treat person re-identification as a special case of the generic instance search problem, where the instance of interest is now a specific person, and address the problem using the same attribute-based approach as for other types of instance search, \eg, shoes and buildings.


\subsection{Contributions}
Our work makes the following contributions. We propose to pursue generic instance search from 1 example where the instance can be an arbitrary 3D-object recorded from a wide range of imaging angles. We argue that this problem is harder than the approximately one-sided instance search of buildings~\cite{philbin2007object}, logos~\cite{joly2009logo} and remote scenes~\cite{jegou2008hamming}. We evaluate state-of-the-art methods on this problem. We observe what works best for buildings loses its generality for shoes and reversely what works worse for buildings may work well for shoes.

Second, we propose to use automatically learned category-specific attributes to handle the wide range of appearance variations in generic instance search. Here we assume we know the category of the query instance which provides critical knowledge when there is only 1 query image. Information of the query category can be given through interactive user interface or automatic image categorization (\eg, shoe, dress, \etc). On the problem of searching among instances from the same category as the query, our category-specific attributes outperform existing instance search methods by a large margin when large appearance variations exist.

Third, inspired by~\cite{zheng2015query}, we treat person re-identification as a special case of generic instance search, where the instance of interest is a specific person. On the popular VIPeR dataset~\cite{gray2007evaluating}, we reach state-of-the-art performance with the same attribute-based method.

As our fourth contribution, we extend our method to search instances without restricting to the known category. 
We propose to augment the category-specific attributes with category-level information which is carried by high-level deep learning features learned from large-scale image categorization and the category-level classification scores. We show that combining category-level information with category-specific attributes achieves superior performance to combining category information with low-level features such as Fisher vector. 



A preliminary version of the paper appeared as~\cite{tao2015attributes}. In this paper, we include several new studies. First, we conduct an empirical study of the parameters of the attribute learning method. We also analyze the impact of the underlying features for attribute learning on the search performance. Using multiple features for learning, which as a whole can better capture the various types of visual properties than individuals, we improve the performance over~\cite{tao2015attributes} substantially. And we treat person re-identification~\cite{gong2014person,bedagkar2014survey,vezzani2013people} as another special case of generic instance search where the query is a specific person with the same attribute-based method. On the popular VIPeR dataset~\cite{gray2007evaluating}, competitive result is achieved, on par with the state-of-the-art. This demonstrates the generic capability of our attribute-based instance search algorithm.   

\section{The difficulty of generic instance search}
The first question we raise in this work is how the state-of-the-art methods perform on generic instance search from 1 example where the query instance can be an arbitrary object. \textit{Can we search for other objects like shoes using the same method that has been shown promising for buildings?} To that end, we evaluate several existing instance search algorithms on both buildings and shoes.

We evaluate the following methods. \textbf{\textit{ExpVLAD}:}~\cite{tao2014locality} introduces locality at two levels to improve instance search from one example. The method considers locality in the picture by evaluating multiple candidate locations in each of the database images. It also considers locality in the feature space by efficiently employing a large visual vocabulary for VLAD and Fisher vector and by an exponential similarity function to give disproportionally high scores on close local descriptor pairs. The locality in the picture was shown effective when searching for instances covering only a part of the image. And the the locality in the feature space was shown useful on all the datasets considered in the reference. 
\textbf{\textit{Triemb}:}~\cite{jegou2014triangulation} proposes triangulation embedding and democratic aggregation. The triangulation embedding encodes a local descriptor with respect to the visual word centers using only directions, not magnitudes. As shown in the paper, the triangulation embedding outperforms Fisher vector~\cite{sanchez2013image}. The democratic aggregation assigns a weight to each local descriptor extracted from an image to ensure all descriptors contribute equally to the self-similarity of the image. This aggregation scheme was shown better than the sum aggregation.  
\textbf{\textit{Fisher}:} We also consider Fisher vector as it has been widely applied in instance search and object categorization where good performance has been reported~\cite{jegou2012aggregating,sanchez2013image}.
\textbf{\textit{Deep-FC}:} It has been shown recently that the activations in the fully connected layers of a deep convolutional neural network (CNN)~\cite{krizhevsky2012imagenet} serve as good features for several computer vision tasks~\cite{razavian2014cnn,babenko2014neural,girshick2014rich}.
\textbf{\textit{VLAD-Conv}:} Very recently,~\cite{ng2015exploiting} proposes to apply VLAD encoding~\cite{jegou2012aggregating} on the output of the convolutional layers of CNN for instance search.

\begin{figure}
    \centering {
        \subfloat{\label{fig:examples_oxford5k}\includegraphics[width=0.9\linewidth]{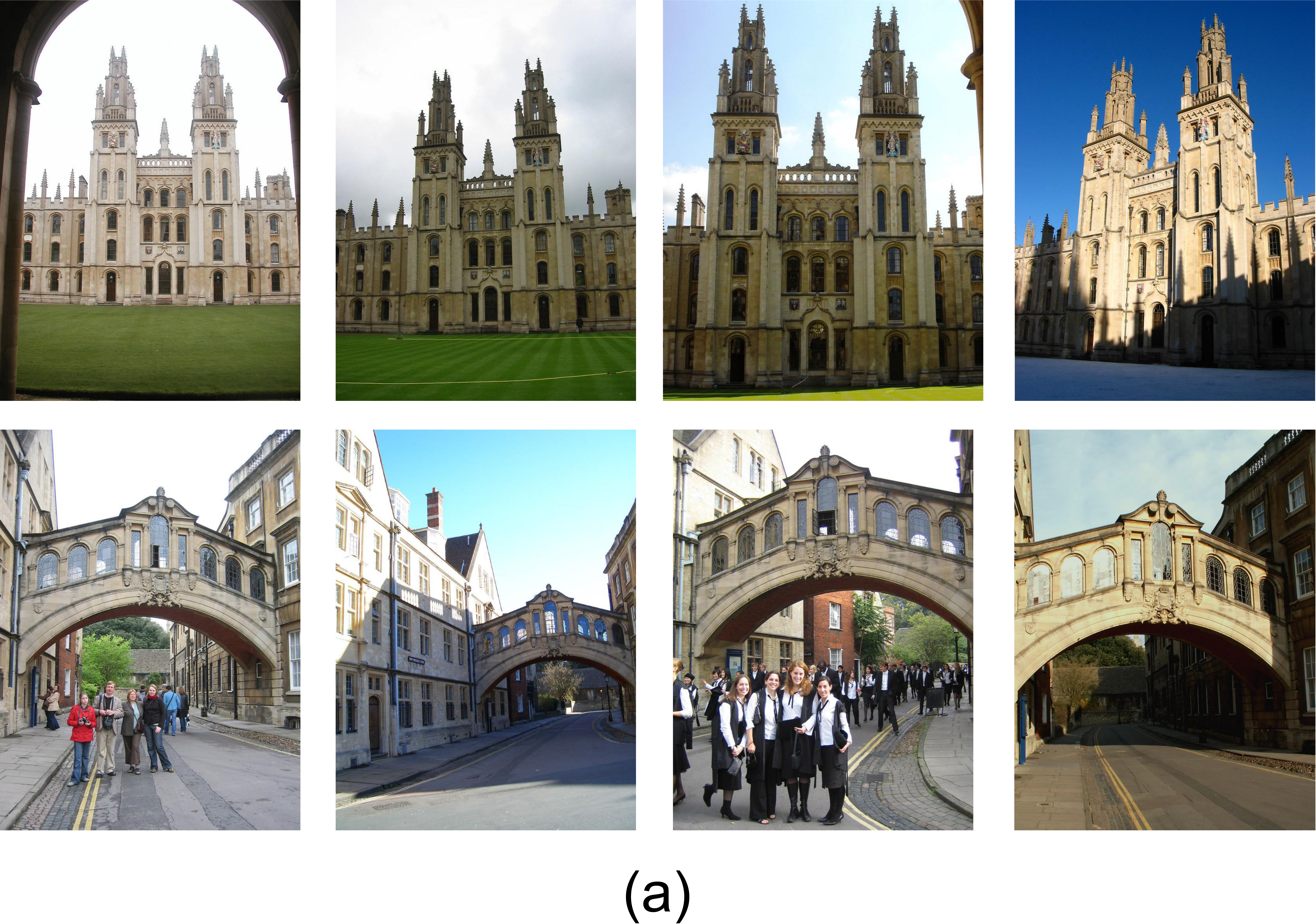}} \\
        \subfloat{\label{fig:example_shoes}\includegraphics[width=0.9\linewidth]{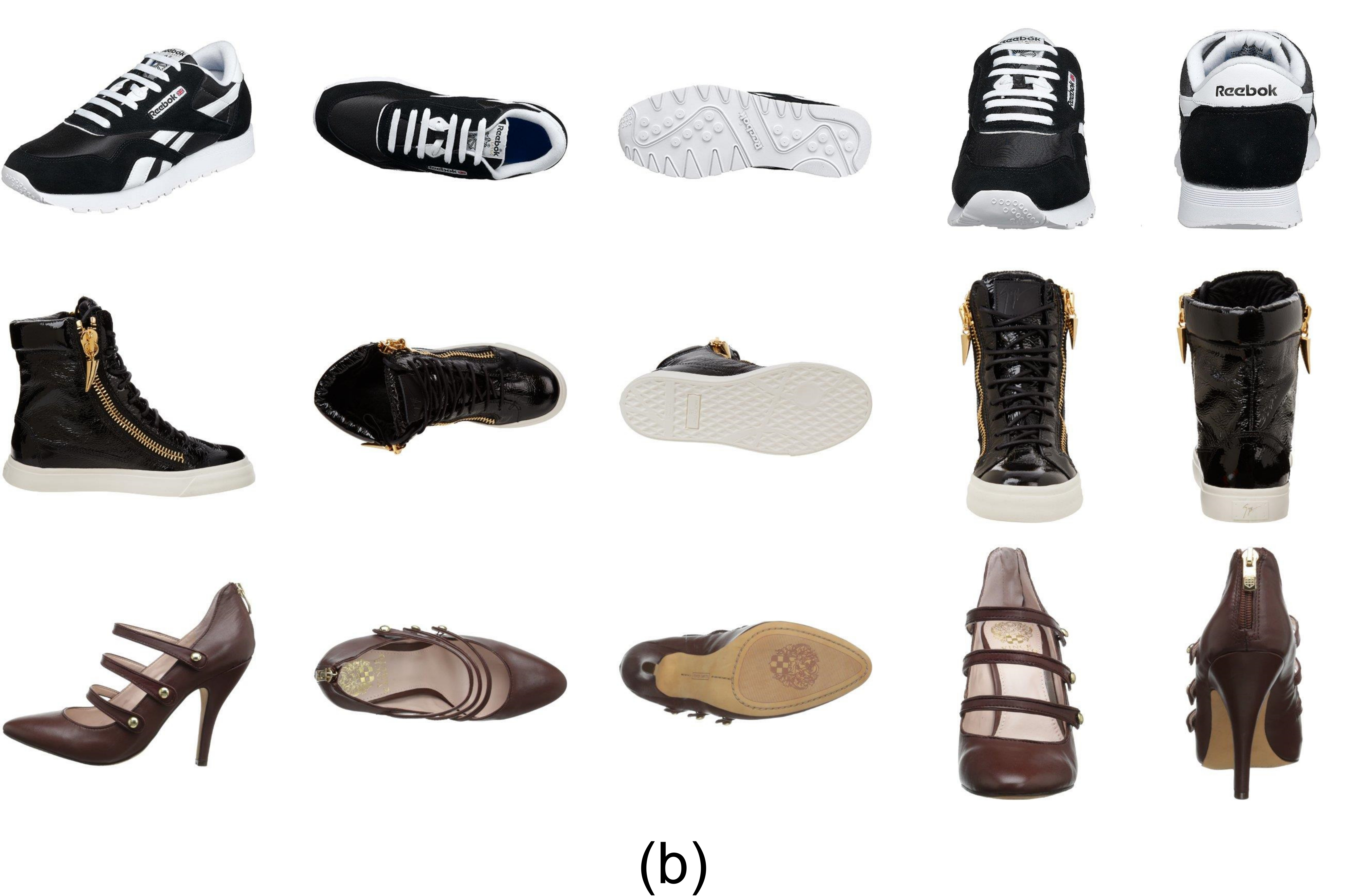}}
    }
    \caption{(a) Examples of two buildings from \textit{Oxford5k}, and (b) Examples of three shoes from our \textit{CleanShoes} dataset. There exists a much wider range of viewpoint variability in the shoe images.}
    \label{fig:examples}
\end{figure}

\textbf{Datasets.} Oxford buildings dataset~\cite{philbin2007object}, often referred to as \textit{Oxford5k}, contains 5062 images downloaded from Flickr. 55 queries of Oxford landmarks are defined, each by a query example. \textit{Oxford5k} is one of the most popular datasets for instance search, which has been used by many works to evaluate their approaches. Figure~\ref{fig:examples_oxford5k} shows examples of two buildings from the dataset.

As a second dataset, we collect a set of shoe images from Amazon\footnote{The properties are with the respective owners. The images are shown here only for scientific purposes.}. It consists of 1000 different shoes and in total 6624 images. Each shoe is recorded from multiple imaging angles including views from front, back, top, bottom, side and some others. One image of a shoe is considered as the query and the goal is to retrieve all the other images of the same shoe. Although these images are with clean background as often seen on shopping websites, this is a challenging dataset mainly due to the presence of considerably large viewpoint variations and self-occlusion. We refer to this dataset as \textit{CleanShoes}. Figure~\ref{fig:example_shoes} shows examples of three shoes from \textit{CleanShoes}. 
There is a shoe dataset available, proposed by~\cite{berg2010automatic}. However, this dataset is not suited for instance search as it does not contain multiple images for one shoe.~\cite{shen2012mobile} also considers shoe images, but the images are well aligned, whereas the images in \textit{CleanShoes} provide a much wider range of viewpoint variations.

\textbf{Implementation details.} For \textit{ExpVLAD}, \textit{Triemb} and \textit{Fisher}, we use the Hessian-Affine detector~\cite{perdoch2009efficient} to extract interest points. The SIFT descriptors are turned into RootSIFT~\cite{arandjelovic2012three}. The full 128D descriptors are used for \textit{ExpVLAD} and \textit{Triemb}, following~\cite{tao2014locality,jegou2014triangulation}, while for \textit{Fisher}, the local descriptor is reduced to 64D using PCA, as the PCA reduction has been shown important for Fisher vector~\cite{jegou2012aggregating,sanchez2013image}. The vocabulary size is 20k, 64 and 256 for \textit{ExpVLAD}, \textit{Triemb} and \textit{Fisher} respectively, following the corresponding references~\cite{tao2014locality,jegou2014triangulation,jegou2012aggregating}. We additionally run a version of Fisher vector with densely sampled RGB-SIFT descriptors~\cite{van2010evaluating} and a vocabulary of 256 components, denoted by \textit{Fisher-D}.
For \textit{Deep-FC}, we use an in-house implementation of the AlexNet~\cite{krizhevsky2012imagenet} trained on ImageNet categories, and take the $\ell_2$ normalized output of the second fully connected layer as the image representation. For \textit{VLAD-Conv}, we apply VLAD encoding with a vocabulary of 100 centers on the conv5\_1 responses of the VGGNet~\cite{simonyan2014very}, following~\cite{ng2015exploiting}. For \textit{Triemb}, \textit{Fisher}, \textit{Fisher-D} and \textit{VLAD-Conv}, power normalization~\cite{perronnin2010improving} and $\ell_2$ normalization are applied.

\begin{figure}
    \centering {
        \includegraphics[width=0.9\linewidth]{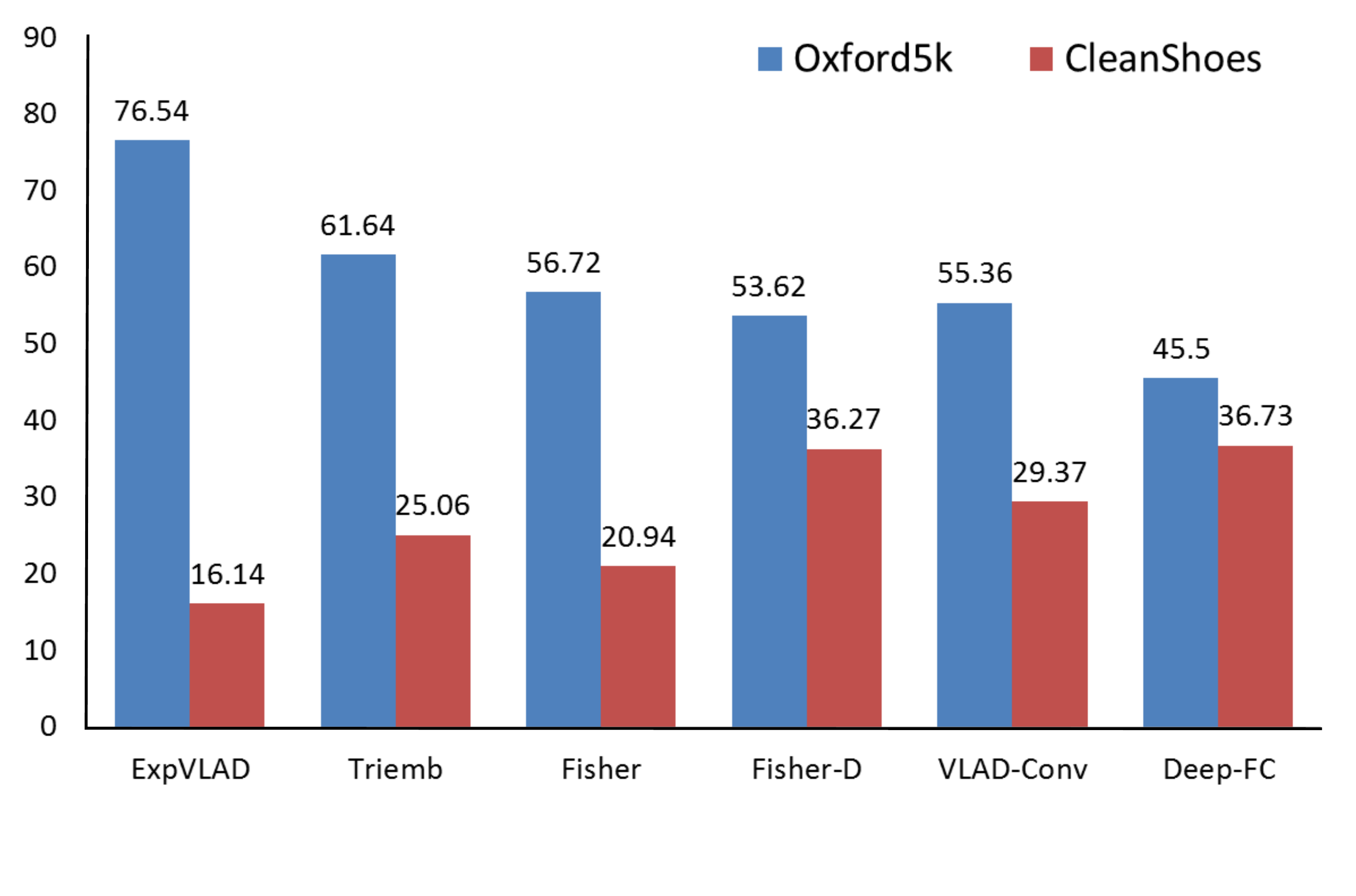}
    }
    \vspace{-5mm}
    \caption{Performance of various state-of-the-art methods for instance search measured in mean average precision\%: ExpVLAD~\cite{tao2014locality}, Triemb~\cite{jegou2014triangulation}, Fisher~\cite{jegou2012aggregating}, VLAD-Conv~\cite{ng2015exploiting} and Deep-FC~\cite{krizhevsky2012imagenet}. For Fisher vector, we consider two versions. \textit{Fisher} denotes the version with interest points and SIFT descriptors, and \textit{Fisher-D} uses densely sampled RGB-SIFT descriptors. 
    \textit{ExpVLAD} achieves better performance than others on \textit{Oxford5k}, but gives lowest result on \textit{CleanShoes}. On the other hand, \textit{Deep-FC} obtains best performance on \textit{CleanShoes}, but has lower result than others on \textit{Oxford5k}.}
    \label{fig:generic_ins_eval}
\end{figure}

\textbf{Results and discussions.} Figure~\ref{fig:generic_ins_eval} summarizes the results on \textit{Oxford5k} and \textit{CleanShoes}. 
\textit{ExpVLAD} adopts a large vocabulary with 20k visual words and the exponential similarity function. As a result, only close local descriptor pairs in the feature space matter in measuring the similarity of two examples. This results in better performance than others on \textit{Oxford5k} where close and relevant local descriptor pairs do exist. However, on the shoe images where close and true matches of local descriptors are rarely present due to the large appearance variations, \textit{ExpVLAD} achieves lowest performance. Both \textit{Triemb} and \textit{Fisher} obtain quite good results on buildings but the results on shoes are low. This is again caused by the fact that local descriptor matching is not reliable on the shoe images where large viewing angle differences are present. \textit{Triemb} outperforms \textit{Fisher}, consistent with the observations in~\cite{jegou2014triangulation}. In this work, we do not consider the RN normalization~\cite{jegou2014triangulation} because it requires extra training data to learn the projection matrix and it does not affect the conclusion we make here. \textit{Fisher-D} works better than \textit{Fisher} on \textit{CleanShoes} by using color information and densely sampled points. Color is a useful cue for discriminating different shoes, and dense sampling is better than interest point detector on shoes which do not have rich textural patterns. However, \textit{Fisher-D} does not improve over \textit{Fisher} on \textit{Oxford5k}. \textit{VLAD-Conv} is in the middle on both sets. \textit{Deep-FC} has lowest performance on buildings, but outperforms others on shoes.

Overall, the performance on shoes is much lower than on the buildings. More interestingly, \textit{ExpVLAD} achieves better performance than others on \textit{Oxford5k}, but gives lowest result on \textit{CleanShoes}. On the other hand, \textit{Deep-FC} obtains best performance on \textit{CleanShoes}, but has lower result than others on \textit{Oxford5k}. We conclude that none of the existing methods work well on both buildings, as an example of 2D one-sided instance search, and shoes, as an example of 3D full-view instance search.

\label{sec::GenericInstanceSearch}

\section{Attributes for generic instance search}\label{sec::attributes}

Attributes, as a higher level abstraction of visual properties, have been shown advantageous in categorization when the training examples are insufficiently covering the appearance variations in the original feature space~\cite{farhadi2009describing,yu2013designing,akata2013label}. In our problem, there is only 1 example available and there still exists a wide range of appearance variations. \textit{Can we employ attributes to address generic instance search?} 

In the literature, two types of attributes have been studied, manually defined attributes with names~\cite{lampert2009learning,akata2013label} and automatically learned unnameable attributes~\cite{yu2013designing,sharmanska2012augmented}. Obtaining manually defined attributes requires a considerable amount of human efforts and sometimes domain expertise, making it hard to scale up to a large number of attributes. Moreover, the manually picked attributes are not necessarily machine-detectable, and not guaranteed to be useful for the task under consideration~\cite{yu2013designing}.   
On the other hand, learned attributes do not need human annotation and have the capacity to be optimized for the task~\cite{yu2013designing,sharmanska2012augmented}. 
For some tasks, like zero-shot learning~\cite{akata2013label} and image retrieval by textual query~\cite{siddiquie2011image}, it is necessary to use human understandable attributes with names. However, in instance search given 1 image query, having attributes with names is not really necessary. In this work, we use automatically learned attributes. Specifically, we focus on searching among instances known to be of the same category in this section using automatically learned category-specific attributes.

Provided with a set of training instances from a certain category, we aim to learn a list of category-specific attributes and use them to perform instance search on new (unseen) instances from the same category. Concretely, given $m$ training images of $n$ objects ($m>n$ as each object has one or multiple examples), the goal is to learn $k$ attribute detectors. In the search phase, the query image and the dataset images are represented by $k$-dimensional attribute detection scores, and the search is performed by comparing the distances in the $k$-dimensional feature space.

Analogous to the class-attribute mapping in attribute-based categorization~\cite{farhadi2009describing,lampert2009learning,akata2013label}, an instance-attribute mapping $A\in\mathbb{R}^{n\times k}$ is designed automatically. The challenge is how to obtain a useful $A$. 
As the goal in instance search is to differentiate different instances, the attributes should be able to make distinctions among the training instances. On the other hand, as the attributes will be used later for instance search on new, unseen instances, the learned attributes need to be able to generalize on unseen instances. To that end, visually similar training instances are encouraged to share attributes. Attributes specific to one training instance are less likely to generalize on unknown instances than those shared by several training instances. And sharing needs to be restricted only among visually similar training instances as the latent common visual patterns among visually dissimilar instances are less likely to be present and detected on new instances even if they can be learned provided with a high dimensional feature space. Besides, to make the best out of the $k$ attributes, it is desirable to have low redundancy among the attributes. Formally, taking the above considerations into account, we design $A$ by  
\begin{equation}
\begin{aligned}
\underset{A}{\text{maximize}}
& & f_{1}(A)+\lambda f_{2}(A)+\gamma f_{3}(A),
\end{aligned}
\label{eqn:eqn_opt_all}
\end{equation}              
where $f_{1}(A)$, $f_{2}(A)$ and $f_{3}(A)$ are defined as follows:
\begin{equation}
\begin{aligned}
f_{1}(A)=\sum_{i,j}^n\|A_{i\cdot}-A_{j\cdot}\|_{2}^{2}, \\
f_{2}(A)=-\sum_{i,j}^nS_{ij}\|A_{i\cdot}-A_{j\cdot}\|_{2}^{2}, \\
f_{3}(A)=-\|A^{T}A-I\|_{F}^{2}.
\end{aligned}
\label{eqn:eqn_opt}
\end{equation}               
$A_{i\cdot}$, the $i$-th row of $A$, is the attribute representation of the $i$-th instance. $f_{1}(A)$ ensures instance separability. $S$ in $f_{2}(A)$ is the visual proximity matrix, where $S_{ij}$ represents visual similarity between instance $i$ and instance $j$, measured \textit{a priori} in certain visual feature space. The similarity between two training instances is computed as the average similarity between the images of the two instances. $f_{2}(A)$ encourages similar attribute representations between visually similar instances, inducing shareable attributes. $f_{3}(A)$ penalizes redundancy between attributes. $\lambda$ and $\gamma$ are two parameters of the objective. Larger $\lambda$ encourages more attribute sharing among visually similar instances and larger $\gamma$ penalizes more on the redundancy in the learned attributes. This formulation was originally proposed in~\cite{yu2013designing} for category recognition. Following~\cite{yu2013designing}, the optimization problem is solved incrementally by obtaining one column of A, \ie, one attribute at each step. Next we briefly describe the optimization procedure.

The objective (Equation~\ref{eqn:eqn_opt_all}) can be rewritten as 
\begin{equation}
\begin{aligned}
\underset{A}{\text{maximize}}
& & Tr(A^{T}PA)-\gamma \|A^{T}A-I\|_{F}^{2},
\end{aligned}
\end{equation}
where $P=Q-\lambda L$. $Q$ is an $n\times n$ matrix with diagonal elements being $n-1$ and off-diagonal elements being $-1$. $L$ is the Laplacian of $S$~\cite{von2007tutorial}. Initializing $A$ as an empty matrix, $A$ can be learned incrementally, one column at one step, by  
\begin{equation}
\begin{aligned}
\underset{\mathbf{a}}{\text{maximize}}
& & \mathbf{a}^{T}R\mathbf{a} & & & s.t. & \mathbf{a}^{T}\mathbf{a}=1,
\end{aligned}
\end{equation}
where $R=P-2\gamma AA^{T}$. The optimal $\mathbf{a}$ is the eigenvector of $R$ with the largest eigenvalue. $A$ is updated by $A=[A, \mathbf{a}]$ at every step. 
In this work, each attribute, \ie, $\mathbf{a}$, is binarized during the optimization. 

\textbf{Attribute detectors.} Once the instance-attribute mapping $A$ has been obtained, the next step is to learn the attribute detectors. In this work, the attribute detectors are formulated as linear SVM classifiers. To train the $j$-th attribute detector, images of the training instances with $A_{ij}>0$ are used as positive examples and the rest images are negative examples\footnote{We have also tried designing the instance-attribute mapping $A$ with continuous values and learning a regressor for each attribute. However, this is not better in terms of instance search performance.}. 

\textbf{Attribute representation.} Given a new image, the attribute representation is generated by applying all the learned attribute detectors and concatenating the SVM classification scores. 
The attribute representation is discriminative in distinguishing different instances as it is optimized to be so when designing $A$. The attribute representation is invariant to the appearance variations of an instance as the invariance is built in the attribute detectors which take all the images of one instance as either all positive or all negative during learning.

\subsection{Experiments}

\subsubsection{Datasets}
\begin{figure}
    \centering {
        \includegraphics[width=0.9\linewidth]{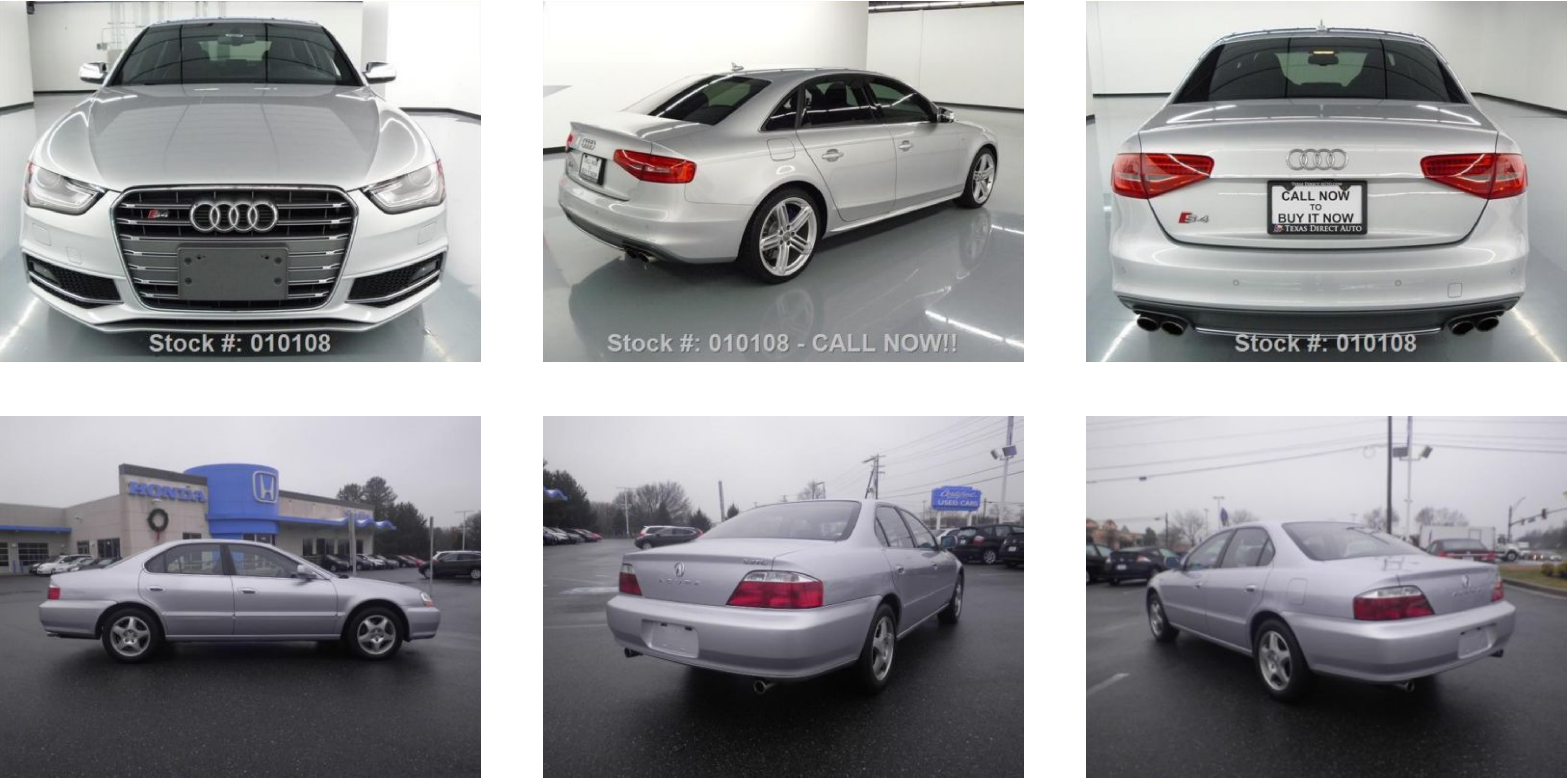}
    }
    \caption{Examples of two cars from the dataset \textit{Cars}.}
    \label{fig:example_cars}
\end{figure}

\textit{Evaluation sets.} The category-specific attributes as learned are evaluated on shoes, cars and buildings. For shoes, the dataset \textit{CleanShoes} described Section~\ref{sec::GenericInstanceSearch} is used. For cars, we collect 1110 images of 270 cars from eBay, denoted by \textit{Cars}. Figure~\ref{fig:example_cars} shows some images of two cars\footnote{The properties are with the respective owners. The images are shown here only for scientific purposes.}. For buildings, a dataset is composed by gathering all 567 images of the 55 Oxford landmarks from \textit{Oxford5k}, denoted by \textit{OxfordPure}. We reuse the 55 queries defined in \textit{Oxford5k}.

\textit{Training sets.} To learn shoe-specific attributes, we collect 2100 images of 300 shoes from Amazon. To train car-specific attributes, we collect 1520 images of 300 cars from eBay. 
To learn building-specific attributes, we use a subset of the large building dataset introduced in~\cite{babenko2014neural}. 
We randomly pick 30 images per class and select automatically the 300 classes that are most relevant to \textit{OxfordPure} according to the visual similarity. We end up with in total 8756 images as some URLs are broken and some classes have less than 30 examples. For all shoes, cars and buildings, the instances in the evaluation sets are not present in the training sets.

\subsubsection{Empirical parameter study} 
We empirically investigate the effect of the two parameters of the learning algorithm ($\lambda$ and $\gamma$ in Equation~\ref{eqn:eqn_opt_all}) on the search performance. We learn different sets of category-specific attributes with different $\lambda$ and $\gamma$ values and evaluate the instance search performance. The study is conducted on the shoe dataset. 

Fisher vector~\cite{sanchez2013image} with densely sampled RGB-SIFT~\cite{van2010evaluating} is used as the underlying representation to compute the visual proximity matrix $S$ in Equation~\ref{eqn:eqn_opt} and learn the attribute detectors. $S$ is built as a mutual 60-NN adjacent matrix throughout the paper. 

\begin{figure}
    \centering {
        \includegraphics[width=0.9\linewidth]{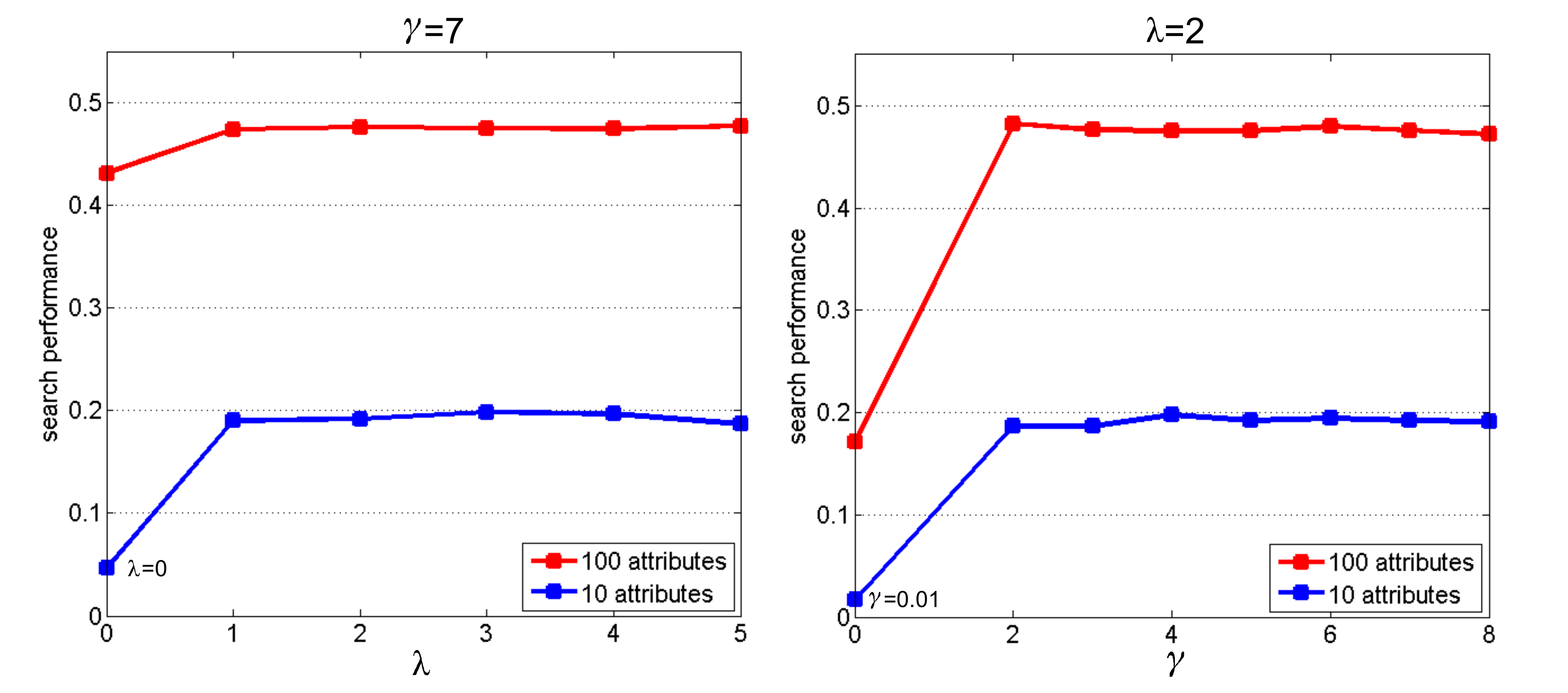}
    }
    \caption{The impact of the parameters of the attribute learning algorithm ($\lambda$ and $\gamma$) on the search performance, measured in mean average precision. The experiments are conducted on \textit{CleanShoes}. When there is no attribute sharing enforced between instances ($\lambda=0$) or there is large redundancy in the learned attributes ($\gamma=0.01$), the search performance is low. It indicates the importance of enforcing attribute sharing and low redundancy. The observation on the impact of $\lambda$ holds when fixing $\gamma$ to other values (The same holds for the observation on $\gamma$.).}
    \label{fig:parameters}
\end{figure}

First, we study the effect of $\lambda$ by fixing $\gamma$. An extreme case is setting $\lambda$ to be 0, which means no attribute sharing among training instances. As shown in Figure~\ref{fig:parameters} (left), when $\lambda$ is 0, the search performance is much worse than when $\lambda$ is from $1$ to $5$, especially when the number of attributes is low. When there is no sharing induced, the learned attributes on the training instances cannot generalize well on the new instances in the search set. As long as sharing is enabled, the search performance is robust to the value of $\lambda$. 

Second, we study the effect of $\gamma$ by fixing $\lambda$. As can be seen from Figure~\ref{fig:parameters} (right), when $\gamma$ is small ($0.01$), which means large redundancy in the learned attributes, the search performance is very low, but stabilizes once $\gamma$ is large enough. 

The above study shows the importance of enforcing attribute sharing and low redundancy during learning as well as the robustness of the learning algorithm against the values of $\lambda$ and $\gamma$, in terms of the instance search performance. 
In the rest of paper, we set $\lambda$ and $\gamma$ to be 2 and 7 respectively to be consistent with the earlier version of the work~\cite{tao2015attributes}.

\subsubsection{Comparison with manual attributes}
We compare the learned attributes with manually defined attributes on shoe search. For manually defined attributes, we use the list of attributes proposed by~\cite{huang2014circle}. We manually annotate the same 2100 training images. In the reference, 42 attributes are defined. However, we merge \textit{super-high} and \textit{high} of ``upper'' and ``heel height'' because it is hard to annotate \textit{super-high} and \textit{high} as two different attributes. This results in 40 attributes. 

\begin{figure}
    \centering {
        \includegraphics[width=0.9\linewidth]{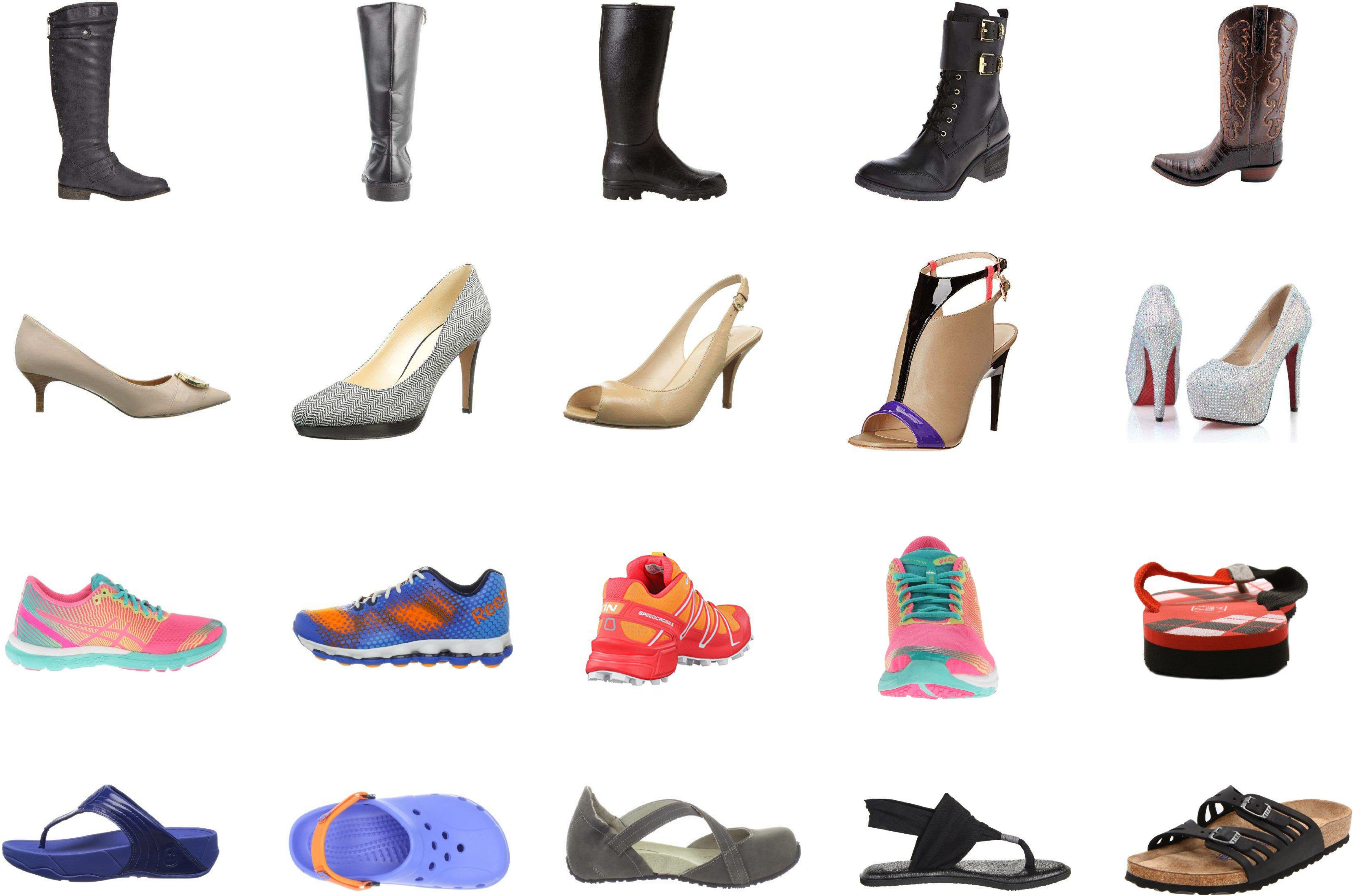}
    }
    \caption{Four automatically designed attributes. Each row is one attribute and the shoes are the ones that have high response for that attribute. Although the automatically learned attributes have no semantic names, apparently they capture sharing patterns among shoes. The first attribute represents high boots. The second describes the high heels. The third is probably about colorfulness. The last one is about openness. The first two are also found in the manually defined attributes while the other two are novel ones discovered automatically.}
    \label{fig:example_attributes}
\end{figure}

Again, Fisher vector is used as the underlying representation to learn attribute detectors. As shown in Table~\ref{tab:generic_ins_attr}, with the same number of attributes, the automatically learned attributes work significantly better than the manual attributes. Moreover, automatically learned attributes are easily scalable, improving performance further. Figure~\ref{fig:example_attributes} shows four automatically learned attributes. Although the attributes have no explicit names, they do capture common visual properties between shoes. 

\begin{table}
    \renewcommand{\arraystretch}{1.3}
    \centering
    \scalebox{1} {
    \setlength{\tabcolsep}{6pt}
    \begin{tabular}{lcc}
        \toprule
             & \textit{number}  & CleanShoes \\
        \midrule
            Manual attributes & \textit{40} & 18.99 \\
            Learned attributes & \textit{40} & 39.44 \\
            Learned attributes & \textit{1000} & 56.57 \\
        \bottomrule
    \end{tabular}
    }
    \caption{Comparison of learned attributes and manually defined attributes on shoe search. The performance is measured in mean average precision\%.} 
    \label{tab:generic_ins_attr}
\end{table}

\subsubsection{Empirical study of underlying feature representation}
In theory, attributes can be learned from any underlying feature representation. In this section, we empirically evaluate the impact of various underlying features for attribute learning on the instance search performance. We consider 5 different feature representations investigated in Section~\ref{sec::GenericInstanceSearch}, \ie, \textit{Triemb}, \textit{Fisher}, \textit{Fisher-D}, \textit{VLAD-Conv} and \textit{Deep-FC}. \textit{ExpVLAD} is not included as it does explicitly form a vector representation to facilitate the learning. The proximity matrix $S$ is measured in the same feature space as used for learning attributes.

First, we evaluate the attributes learned from single underlying features and compare with existing approaches.
The results are summarized in Table~\ref{tab:generic_ins_comp}. We observe that when the underlying feature representation for attribute learning is based on sparse interest points, including \textit{Triemb} and \textit{Fisher}, the learned attribute representation does not always improve the search performance over the original representation. However, when the underlying feature representation is based on densely extracted visual cues, including \textit{Fisher-D}, \textit{VLAD-Conv} and \textit{Deep-FC}, the attribute representation always outperforms the underlying feature representation by a large margin. This indicates that the mapping from the original feature representation to the attribute representation is selective. It selects the useful information which is discriminative among different instances and invariant to the variations of the same instance, while discarding other disturbing information. We argue that a large amount of useful information has already been filtered by the internal selection step of the interest point detector and therefore attribute representation learned on interest point based features does not help much. 
The attribute representation learned using \textit{VLAD-Conv} achieves better performance than those learned from other underlying representations. On the shoe and car datasets, the learned attribute representation significantly outperforms existing approaches. Attributes are superior in addressing the large appearance variations caused by the large imaging angle difference present in the shoe and car images, even though the attributes are learned from other instances. The attribute representation also works well on the buildings. In addition, attribute representation has a much lower dimensionality than other representations.

\begin{table}
    \renewcommand{\arraystretch}{1.3}
    \centering
    \scalebox{0.8} {
    \setlength{\tabcolsep}{6pt}
    \begin{tabular}{lcccc}
        \toprule
             & \textit{dim} & CleanShoes & Cars & OxfordPure \\
        \midrule
             ExpVLAD~\cite{tao2014locality} & \textemdash & 16.14 & 23.70 & \textbf{87.01} \\
             Triemb~\cite{jegou2014triangulation} & \textit{8064} & 25.06 & 18.56 & 75.33 \\
             Fisher~\cite{jegou2012aggregating} & \textit{16384} & 20.94 & 18.37 & 70.81 \\ 
             Fisher-D~\cite{jegou2012aggregating,van2010evaluating} & \textit{40960} & 36.27 & 20.89 & 67.41 \\
             VLAD-Conv~\cite{ng2015exploiting} & \textit{51200} & 29.37 & 27.27 & 69.05 \\
             Deep-FC~\cite{krizhevsky2012imagenet} & \textit{4096} & 36.73 & 22.36 & 59.48 \\
        \midrule
             Attributes(Triemb) & \textit{1000} & 19.83 & 28.15 & 71.58 \\
             Attributes(Fisher) & \textit{1000} & 17.67 & 31.21 & 69.33 \\
             Attributes(Fisher-D) & \textit{1000} & 56.57 & 51.11 & 77.36 \\
             Attributes(VLAD-Conv) & \textit{1000} & \textbf{63.19} & \textbf{63.99} & 82.86 \\
             Attributes(Deep-FC) & \textit{1000} & 57.11 & 38.07 & 69.51 \\
             
        \bottomrule
    \end{tabular}
    }
    \caption{Performance in mean average precision\% of existing methods (top part of the table) and the attributes learned from single underlying features (bottom part). The attributes learned from \textit{Fisher-D}, \textit{VLAD-Conv} or \textit{Deep-FC} outperform existing methods significantly on shoes and cars, and achieve comparable performance on buildings. Attributes learned from the underlying features that capture densely the visual cues (\textit{Fisher-D}, \textit{VLAD-Conv} and \textit{Deep-FC}) are better than those learned from the underlying features based on sparse interest points (\textit{Triemb} and \textit{Fisher}).}
    \label{tab:generic_ins_comp}
\end{table} 

\begin{table*}
    \renewcommand{\arraystretch}{1.3}
    \centering
    \scalebox{0.85} {
    \setlength{\tabcolsep}{6pt}
    \begin{tabular}{lccccc}
        \toprule
             &  \textit{dim} & CleanShoes & Cars & OxfordPure \\
        \midrule
             Fisher-D~\cite{jegou2012aggregating,van2010evaluating},VLAD-Conv~\cite{ng2015exploiting} & \textit{92160} & 35.64 & 26.18  & 71.29  \\
             Fisher-D~\cite{jegou2012aggregating,van2010evaluating},Deep-FC~\cite{krizhevsky2012imagenet} & \textit{45056} & 41.55 & 22.65 & 69.41 \\
             VLAD-Conv~\cite{ng2015exploiting},Deep-FC~\cite{krizhevsky2012imagenet} & \textit{55296} & 36.25 & 26.25  & 69.84 \\
             Fisher-D~\cite{jegou2012aggregating,van2010evaluating},VLAD-Conv~\cite{ng2015exploiting},Deep-FC~\cite{krizhevsky2012imagenet} & \textit{96256} & 39.04 & 25.58  & 71.42 \\
        \midrule     
             Attributes(Fisher-D,VLAD-Conv) & \textit{1000} & 63.97 & 69.19  & 83.22  \\
             Attributes(Fisher-D,Deep-FC) & \textit{1000} & 67.45  & 59.96  &  78.66  \\
             Attributes(VLAD-Conv,Deep-FC) & \textit{1000} & 67.06 & 69.02  & \textbf{83.75} \\
             Attributes(Fisher-D,VLAD-Conv,Deep-FC)  & \textit{1000} & \textbf{67.87}  & \textbf{71.74} & 83.06 \\
             
        \bottomrule
    \end{tabular}
    }
    \caption{Performance in mean average precision\% of combining multiple existing representations (top part of the table) and the attributes learned from multiple underlying features (bottom part). The learned attribute representation significantly outperforms the underlying representation. Comparison with Table~\ref{tab:generic_ins_comp} shows that the attribute representation learned from multiple underlying features outperforms those learned on single features. Interestingly, combining the same underlying features and directly using them for instance search without attributes does not necessarily improve over individual features.}
    \label{tab:attr_multifeats}
\end{table*} 

Second, in Table~\ref{tab:attr_multifeats}, we investigate the effects of using multiple underlying features for attribute learning. Again, the attribute representation outperforms the underlying feature representation significantly.
Comparing Table~\ref{tab:attr_multifeats} and Table~\ref{tab:generic_ins_comp}, it is clear that the attribute representation learned on the combination of multiple underlying features outperforms those learned on single features. This demonstrates the advantage of using multiple underlying feature representations, which as a whole can better capture the various types of visual properties than a single representation. Interestingly, combining the same underlying features and directly using them for instance search without attributes does not necessarily improve over individual features, which confirms again the advantage of attributes. The attribute representation learned on the combination of \textit{Fisher-D}, \textit{VLAD-Conv} and \textit{Deep-FC} achieves best performance on shoes and cars, and close to best performance on buildings, improving the results reported in the earlier version of the work~\cite{tao2015attributes} from 56.57\% to 67.87\% on \textit{CleanShoes}, from 51.11\% to 71.74\% on \textit{Cars}, and from 77.36\% to 83.06\% on \textit{OxfordPure} in mean average precision.

\section{Person re-identification as instance search}
Person re-identification is the problem of identifying the images in a database which depict the same person as in the probe image. The probe image and the relevant images in the database are usually captured by different cameras with different recording settings, causing large viewpoint and illumination variations. Besides, a person might have different poses in different recordings and might be partially occluded. All these result in large intra-person variations, making person re-identification a challenging problem. 
In this work, we treat person re-identification as a specific person search problem, and address the problem using the attribute-based method presented in Section~\ref{sec::attributes}.


\textbf{Dataset and evaluation protocol.} We use the VIPeR dataset~\cite{gray2007evaluating}. It has been widely used for benchmark evaluation. It contains 632 pedestrians, each recorded by two cameras. One view is considered as the probe image and the goal is to identify the other view of the same person. 
The 632 pairs are randomly divided into two halves, one for training and one for testing. The performance is evaluated using the Cumulative Match Characteristic (CMC) curve~\cite{gray2007evaluating} which estimates the expectation of finding the correct answer in the top $k$ results. The experiment is repeated 10 times to report an average performance\footnote{We use the 10 divisions provided by~\cite{zheng2015query}}.

\textbf{Implementation details.} 1000 attributes detectors are learned using the training split. To learn the attributes, we employ multiple underlying features. We use the bag-of-word histograms on local color histograms (\textit{CH}), local color naming descriptors (\textit{CN}), local HOG (\textit{HOG}) and local LBP descriptor (\textit{LBP}), provided by~\cite{zheng2015query}\footnote{\url{http://www.liangzheng.com.cn/Project/project_fusion.html}}. Besides, we employ \textit{Deep-FC}, \textit{Fisher-D} and \textit{VLAD-Conv}. Vocabularies with 16 components and 8 centers are used for \textit{Fisher-D} and \textit{VLAD-Conv} respectively. The visual proximity matrix $S$ in equation~\ref{eqn:eqn_opt} is built as a mutual 60-NN adjacent matrix, the same as in previous sections.

\begin{figure}
    \centering {
        \includegraphics[width=0.7\linewidth]{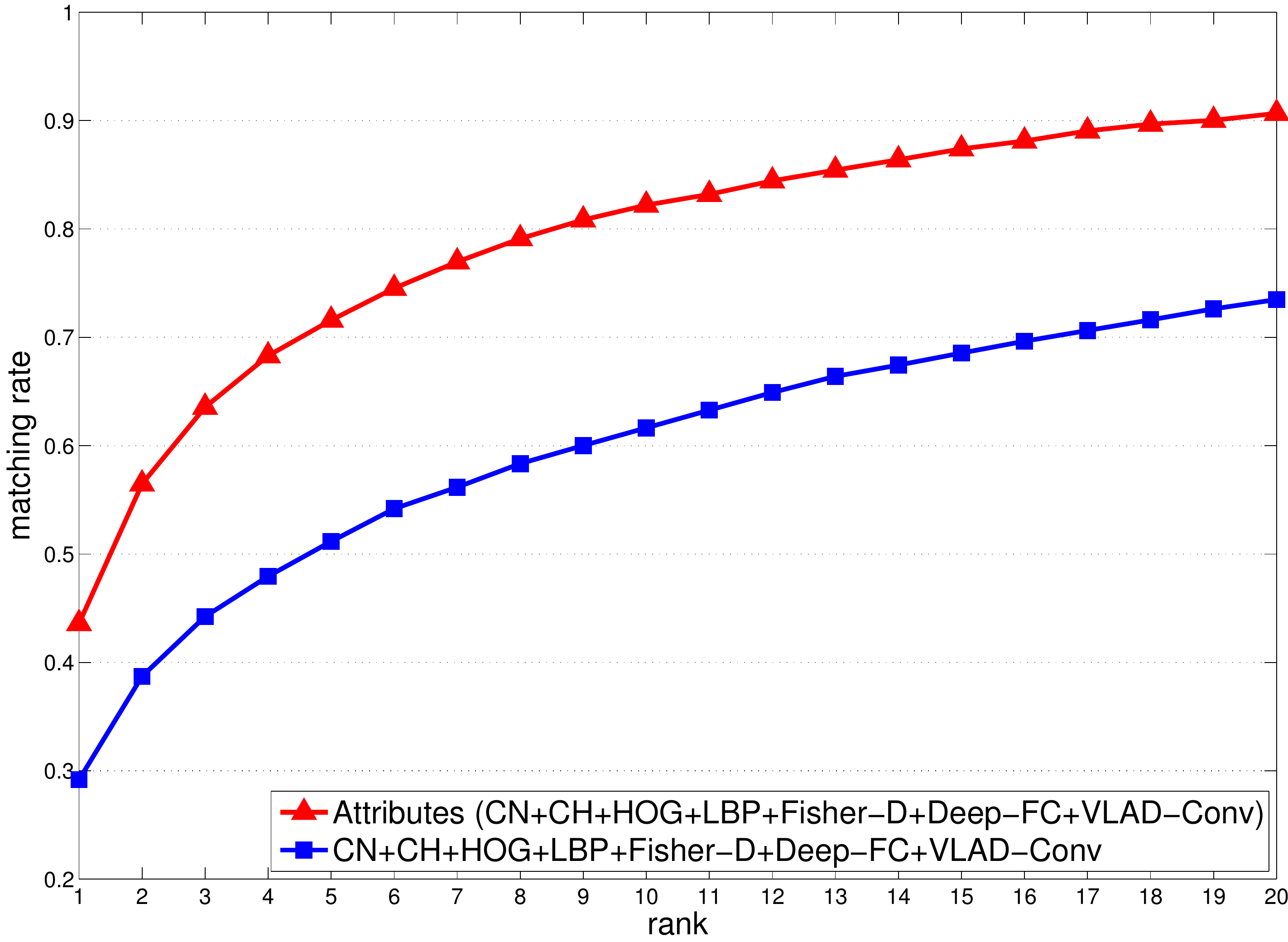}
    }
    \caption{Performance on VIReR dataset~\cite{gray2007evaluating}, measured in matching rate on top ranked images. The learned attribute representation significantly outperforms the original underlying representation.}
    \label{fig:re-id_results_final}
\end{figure}

\begin{table}
    \renewcommand{\arraystretch}{1.3}
    \centering
    \scalebox{0.85} {
    \setlength{\tabcolsep}{6pt}
    \begin{tabular}{lcccc}
        \toprule
             & rank=1 & rank=5 & rank=10 & rank=20 \\
        \midrule
			Zheng \etal~\cite{zheng2015query} & 30.2 & 51.6 & 62.4 & 73.8\\
			Ahmed \etal~\cite{ahmed2015improved} & 34.8 & \textemdash & \textemdash & \textemdash \\
			Chen \etal~\cite{chen2015similarity} & 36.8 & 70.4 & \textbf{83.7} & 91.7 \\
			Shi \etal~\cite{shi2015transferring} & 31.1 & 68.6 & 82.8 & \textbf{94.9} \\
			Liao \etal~\cite{liao2015person} & 40.0 & \textemdash & 80.5 & 91.1 \\
			Paisitkriangkrai \etal~\cite{paisitkriangkrai2015learning} & \textbf{45.9} & \textemdash & \textemdash & \textemdash \\  
        \midrule           
			Ours & 43.6 & \textbf{71.6} & 82.2 & 90.7 \\
        \bottomrule
    \end{tabular}
    }
    \caption{Comparison with state-of-the-art on VIPeR dataset~\cite{gray2007evaluating} by correct matching rates(\%). Although not being specialized for person, our method keeps up with the state-of-the-art for all ranks.}
    \label{tab:re-id_comp}
\end{table} 

\textbf{Results.} As shown in Figure~\ref{fig:re-id_results_final}, the learned attribute representation significantly outperforms the original underlying representation. The learned attributes can handle well the large appearance variations. Table~\ref{tab:re-id_comp} summarizes the comparison with the state-of-the-art. Although the proposed attribute-based method is not specially designed for person re-identification, it achieves good performance, on par with the state-of-the-art. 

\section{Categories and attributes for generic instance search}
In this section, we consider searching for an instance from a dataset which contains instances from various categories. 
As the category-specific attributes are optimized to make distinctions among instances of the same category, they might not be able to distinguish the instance of interest from the instances of other categories. In order to address the possible confusion of the query instance with instances from other categories, we propose to use the category-level information also. 

Ideally one could first categorize all the images in the database and then search using category-specific attributes among the images from the same category as the query. However, as errors made in categorization are irreversible, we choose to avoid explicit binary classification but augment the attributes with category-level information.

We consider two ways to capture the category-level information. First, we adopt the 4096-dimensional output of the second fully connected layer of a CNN~\cite{krizhevsky2012imagenet} as an additional feature, as it has been shown the activations of the top layers of a CNN capture high-level category-related information~\cite{zeiler2014visualizing}. The CNN is trained using ImageNet categories. Second, we build a general category classifier to alleviate the potential problem of the deep learning feature, namely the deep learning feature may bring examples that have common elements with the query instance even if they are irrelevant, such as skins for shoes. 
Combining the two types of category-level information with the category-specific attributes, the similarity between a query $q$ and an example $d$ in the search set is computed by
\begin{equation}
S(q,d)=S_{deep}(q,d)+S_{class}(d)+S_{attr}(q,d),
\end{equation}    
where $S_{deep}(q,d)$ is the similarity of $q$ and $d$ in the deep learning feature space, $S_{class}(d)$ is the classification response on $d$ and $S_{attr}(q,d)$ is the similarity in the attribute space. The three scores are normalized to be $[0,1]$. 

\begin{figure}
    \centering {
        \includegraphics[width=0.95\linewidth]{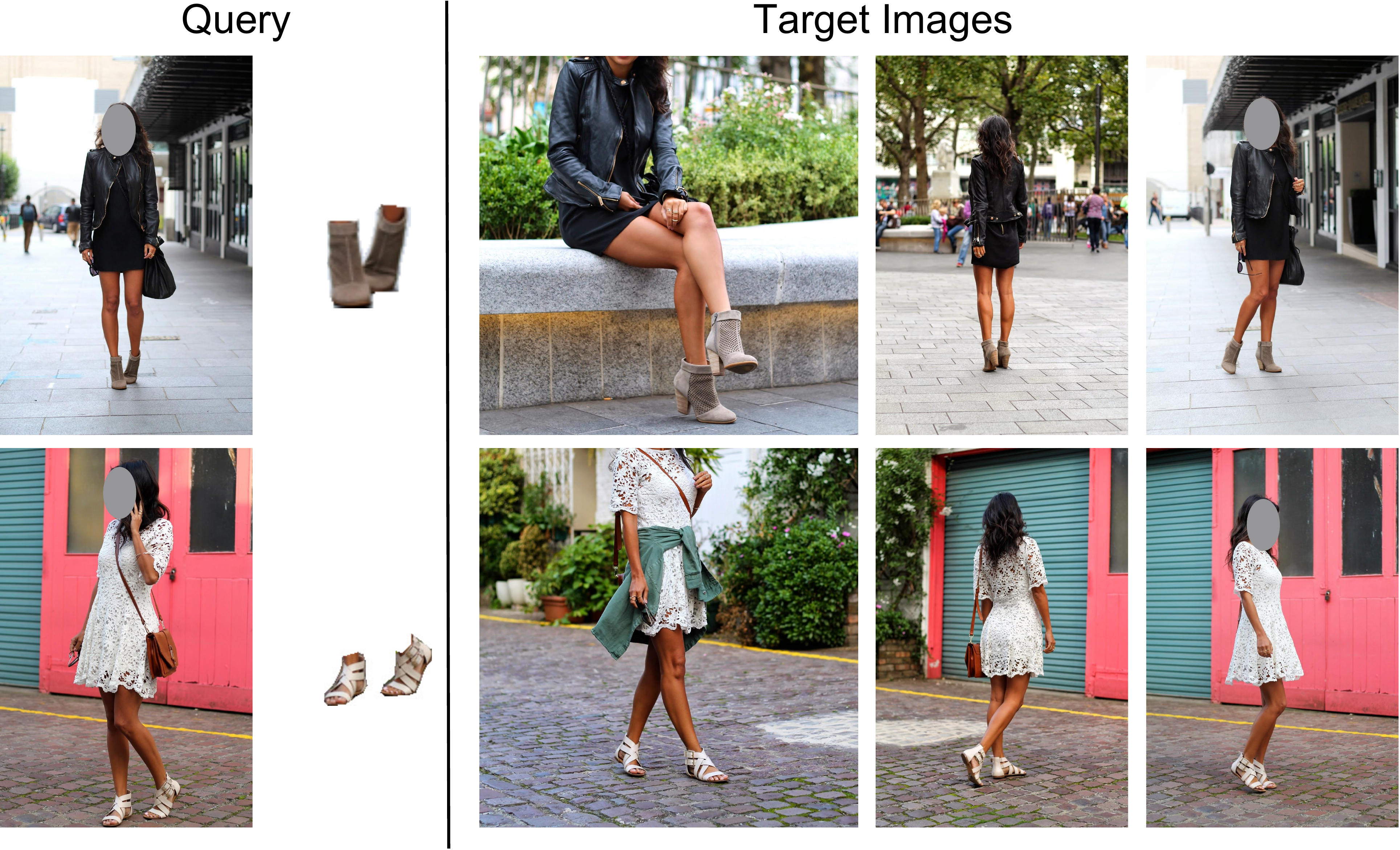}
    }
    \caption{Examples of two shoes from \textit{StreetShoes}. As there is only 1 query example, by manual annotation, we only consider the object region to ensure the object to search is clear, as shown in the second column. The goal is to retrieve from an image collection the target images which depict the same shoe. Note large differences in scale and viewpoint between query and target images.} 
    \label{fig:example_streetshoes}
\end{figure}

\textbf{Datasets.} We evaluate on shoes. A set of 15 shoes and in total 59 images is collected from two fashion blogs\footnote{http://www.pursuitofshoes.com/ and http://www.seaofshoes.com/. The properties are with the respective owners. The images are shown here only for scientific purposes.}. These images are recorded in streets with cluttered background, different from the `clean' images in \textit{CleanShoes}.  
We consider one image of a shoe as the query and aim to find other images of the same shoe. The shoe images are inserted into the test and validation parts of the Pascal VOC 2007 classification dataset~\cite{pascal-voc-2007}. The Pascal dataset provides distractor images. We refer to the dataset containing the shoe images plus distractors as \textit{StreetShoes}. Figure~\ref{fig:example_streetshoes} shows two examples. To learn the shoe classifier, we use the 300 `clean' shoes for attributes learning in Section~\ref{sec::attributes} as positive examples and consider the training part of the Pascal VOC 2007 classification dataset as negative examples.

\begin{table}
    \renewcommand{\arraystretch}{1.3}
    \centering
    \scalebox{0.95} {
    \setlength{\tabcolsep}{6pt}
    \begin{tabular}{lc}
        \toprule
            & StreetShoes\\
        \midrule
          Deep(128D) & 21.68\\
          Fisher(128D) & 9.38 \\
          Attributes(128D) & 3.10 \\
        \midrule
          Deep + Fisher & 19.76 \\
          Deep + Attributes & 18.43 \\
        \midrule
          Deep + Classifier + Fisher & 22.70 \\
          Deep + Classifier + Attributes & \textbf{30.45}\\
        \bottomrule
    \end{tabular}
    }
    \caption{Performance in mean average precision\% on \textit{StreetShoes}. The proposed method of combining the category-specific attributes with two types of category-level information outperforms the combination of category-level information with Fisher vector.} 
    \label{tab:street_shoes}
\end{table} 

\textbf{Implementation details.} As there is 1 query image, by manually annotation we only consider the object region to ensure the target is clear. It is worthwhile to mention that although only the object part in the query image is considered, we cannot completely get rid of skins for some shoes, as shown in Figure~\ref{fig:example_streetshoes}. We use selective search~\cite{uijlings2013selective} to generate many candidate locations in each database image and search over these local objects in the images as~\cite{tao2014locality}. We adopt a short representation with 128 dimensions. Specifically, we reduce the dimensionality of the deep learning features and the attribute representations with a PCA reduction. And for Fisher vectors, we adopt the whitening technique proposed in~\cite{jegou2012negative}, proven better than PCA. We reuse the attribute detectors from Section~\ref{sec::attributes}.

\begin{figure*}
    \centering {
        \includegraphics[width=1\linewidth]{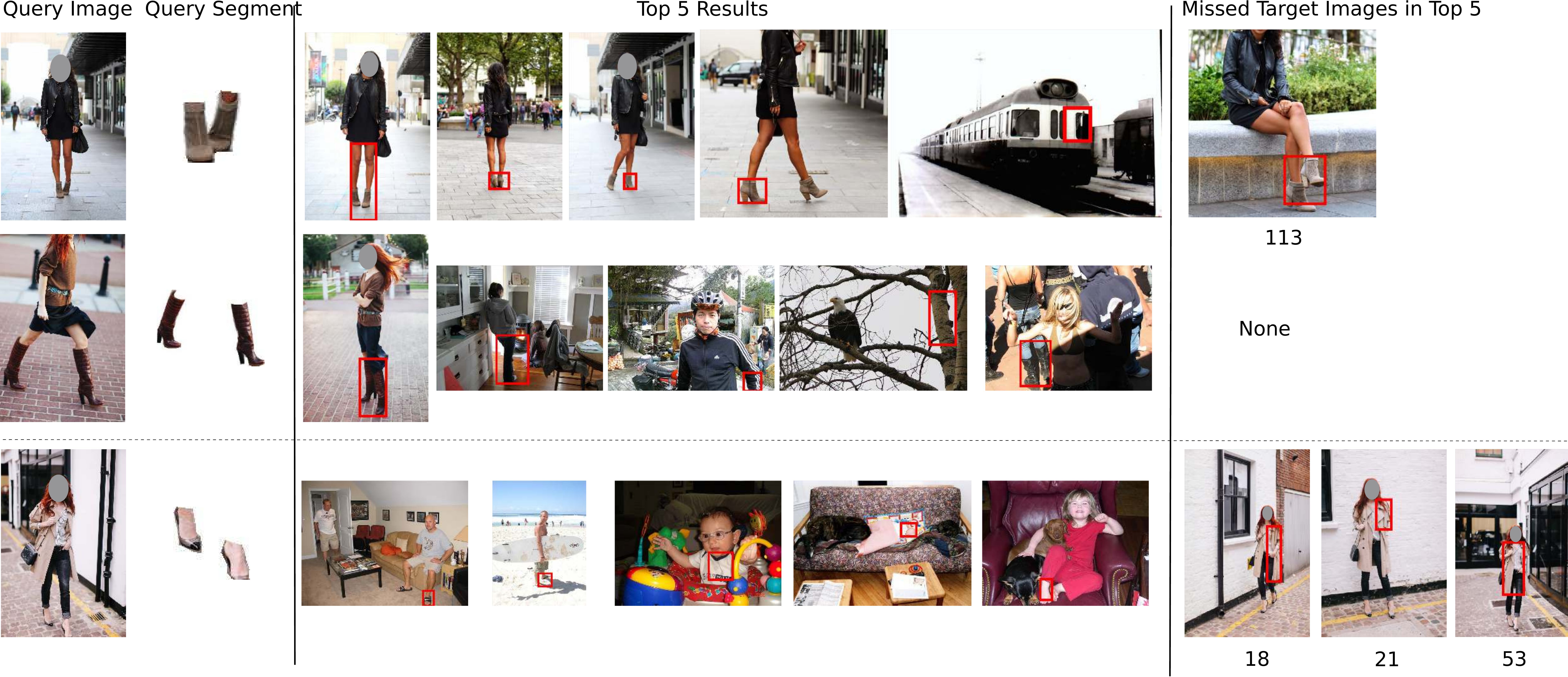}
    }
    \caption{Search results of three query instances, two success cases (the first two) and a failure case (the third one). Only the segment is used as query. For the first instance, it has 5 relevant images in the search set, and 4 of them are returned in the top 5 positions.  For the second instance, there is only 1 relevant example in the search set and it is returned at the first position. For the instance at the bottom, it has 3 relevant images and none of them are returned in the top 5. It is a very hard case, as the shoe is partially visible and the majority of the query segment is about the bare feet. Images of bare footed people appear in the top results. The correct images are ranked at 18, 21 and 53, and they are actually retrieved based on wrong information.}
    \label{fig:results_top5}
\end{figure*}

\textbf{Results and discussions.} The results are shown in Table~\ref{tab:street_shoes}. On \textit{StreetShoes}, the proposed method of combining category-specific attributes with two types of category-level information achieves the best performance, $30.45\%$ in mean average precision. We observe that when considering deep features alone as the category-level information, the system brings many examples of skins. The shoe classifier trained on clean shoe images help eliminate these irrelevant examples. We conclude that the proposed method of combining the category-specific attributes with two types of category-level information is effective, outperforming the combination of category-level information with Fisher vector. Figure~\ref{fig:results_top5} shows the search results of three query instances returned by the proposed method, two success cases and a failure case.

\section{Conclusion}
In this paper, we pursue generic instance search from 1 example. Firstly, we evaluate existing instance search approaches on the problem of generic instance search, illustrated on buildings and shoes, two contrasting categories of objects. We observe that what works for buildings does not necessarily work for shoes. For instance,~\cite{tao2014locality} employs large visual vocabularies and the exponential similarity function to emphasize close matches of local descriptors, resulting in large improvement over other methods when searching for buildings. However, the same approach achieves worst performance when searching for shoes. The reason is that for shoes which have much wider range of viewpoint variability and more sides than buildings, matching local descriptors precisely between two images is not reliable. 

Secondly, we propose to use category-specific attributes to handle the large appearance variations present in generic instance search. We assume the category of the query is known, \eg, from the user input. 
When searching among instances from the same category as the query, attributes outperform existing approaches by a large margin on shoes and cars at the expense of knowing the category of the instance and learning the attributes. For instance search from only one example, it may be reasonable to use more user input. On the building set, the category-specific attributes obtain a comparable performance. 

Thirdly, we consider person re-identification as a special case of generic instance search where the query is a specific person. We show the same attribute-based approach achieves competitive performance, on par with the state-of-the-art in person re-identification.  


Fourthly, we consider searching for an instance in datasets containing instances from various categories. We propose to use the category-level information to address the possible confusion of the query instance with instances from other categories. We show that combining category-level information carried by deep learning features and the categorization scores with the learned category-specific attributes outperforms combining the category information with Fisher vector.

Going back to the experiments using attributes alone, the proposed same method achieves 67.87\% in mean average precision (mAP) on \textit{CleanShoes} for shoe search (Table~\ref{tab:attr_multifeats}), 71.74\% in mAP on \textit{Cars} for car search (Table~\ref{tab:attr_multifeats}), 83.06\% in mAP on \textit{OxfordPure} for building search (Table~\ref{tab:attr_multifeats}) and 43.6\% in matching rate at rank 1 on \textit{VIPeR} for person search (Table~\ref{tab:re-id_comp}), while the best performance of existing methods are 36.73\% (Table~\ref{tab:generic_ins_comp}), 27.27\% (Table~\ref{tab:generic_ins_comp}), 87.01\% (Table~\ref{tab:generic_ins_comp}) and 45.9\% (Table~\ref{tab:re-id_comp}) respectively. The method is generic for instance search indeed. 

\section*{Acknowledgments} 
This research is supported by the Dutch national program COMMIT/.


\bibliography{main}

\begin{thebibliography}{10}
\expandafter\ifx\csname url\endcsname\relax
  \def\url#1{\texttt{#1}}\fi
\expandafter\ifx\csname urlprefix\endcsname\relax\def\urlprefix{URL }\fi
\expandafter\ifx\csname href\endcsname\relax
  \def\href#1#2{#2} \def\path#1{#1}\fi

\bibitem{arandjelovic2012multiple}
R.~Arandjelovic, A.~Zisserman, Multiple queries for large scale specific object
  retrieval., in: British Machine Vision Conference, 2012.

\bibitem{jegou2008hamming}
H.~J{\'e}gou, M.~Douze, C.~Schmid, Hamming embedding and weak geometric
  consistency for large scale image search, in: European Conference on Computer
  Vision, 2008.

\bibitem{over2012trecvid}
P.~Over, G.~Awad, J.~Fiscus, G.~Sanders, B.~Shaw, Trecvid 2012 - an
  introduction of the goals, tasks, data, evaluation mechanisms and metrics,
  in: The TREC Video Retrieval Evaluation (TRECVID), 2012.

\bibitem{qin2011hello}
D.~Qin, S.~Gammeter, L.~Bossard, T.~Quack, L.~Van~Gool, Hello neighbor:
  accurate object retrieval with k-reciprocal nearest neighbors, in: IEEE
  Conference on Computer Vision and Pattern Recognition, 2011.

\bibitem{zhu2013query}
C.-Z. Zhu, H.~J{\'e}gou, S.~Satoh, Query-adaptive asymmetrical dissimilarities
  for visual object retrieval, in: International Conference on Computer Vision,
  2013.

\bibitem{arandjelovic2012three}
R.~Arandjelovi{\'c}, A.~Zisserman, Three things everyone should know to improve
  object retrieval, in: IEEE Conference on Computer Vision and Pattern
  Recognition, 2012.

\bibitem{philbin2007object}
J.~Philbin, O.~Chum, M.~Isard, J.~Sivic, A.~Zisserman, Object retrieval with
  large vocabularies and fast spatial matching, in: IEEE Conference on Computer
  Vision and Pattern Recognition, 2007.

\bibitem{gavves2012visual}
E.~Gavves, C.~G. Snoek, A.~W.~M. Smeulders, Visual synonyms for landmark image
  retrieval, Computer Vision and Image Understanding 116~(2) (2012) 238--249.

\bibitem{joly2009logo}
A.~Joly, O.~Buisson, Logo retrieval with a contrario visual query expansion,
  in: ACM Multimedia Conference, 2009.

\bibitem{revaud2012correlation}
J.~Revaud, M.~Douze, C.~Schmid, Correlation-based burstiness for logo
  retrieval, in: ACM Multimedia Conference, 2012.

\bibitem{tao2014locality}
R.~Tao, E.~Gavves, C.~G.~M. Snoek, A.~W.~M. Smeulders, Locality in generic
  instance search from one example, in: IEEE Conference on Computer Vision and
  Pattern Recognition, 2014.

\bibitem{wang2011contextual}
X.~Wang, M.~Yang, T.~Cour, S.~Zhu, K.~Yu, T.~X. Han, Contextual weighting for
  vocabulary tree based image retrieval, in: International Conference on
  Computer Vision, 2011.

\bibitem{gong2014person}
S.~Gong, M.~Cristani, S.~Yan, C.~C. Loy, Person re-identification, Vol.~1,
  Springer, 2014.

\bibitem{lowe2004distinctive}
D.~G. Lowe, Distinctive image features from scale-invariant keypoints,
  International Journal of Computer Vision 60~(2) (2004) 91--110.

\bibitem{bay2008speeded}
H.~Bay, A.~Ess, T.~Tuytelaars, L.~Van~Gool, Speeded-up robust features (surf),
  Computer Vision and Image Understanding 110~(3) (2008) 346--359.

\bibitem{jegou2014triangulation}
H.~J{\'e}gou, A.~Zisserman, Triangulation embedding and democratic aggregation
  for image search, in: IEEE Conference on Computer Vision and Pattern
  Recognition, 2014.

\bibitem{tolias2013aggregate}
G.~Tolias, Y.~Avrithis, H.~J{\'e}gou, To aggregate or not to aggregate:
  Selective match kernels for image search, in: International Conference on
  Computer Vision, 2013.

\bibitem{mikolajczyk2005performance}
K.~Mikolajczyk, C.~Schmid, A performance evaluation of local descriptors, IEEE
  Transactions on Pattern Analysis and Machine Intelligence 27~(10) (2005)
  1615--1630.

\bibitem{farhadi2009describing}
A.~Farhadi, I.~Endres, D.~Hoiem, D.~Forsyth, Describing objects by their
  attributes, in: IEEE Conference on Computer Vision and Pattern Recognition,
  2009.

\bibitem{lampert2009learning}
C.~H. Lampert, H.~Nickisch, S.~Harmeling, Learning to detect unseen object
  classes by between-class attribute transfer, in: IEEE Conference on Computer
  Vision and Pattern Recognition, 2009.

\bibitem{yu2013designing}
F.~X. Yu, L.~Cao, R.~S. Feris, J.~R. Smith, S.-F. Chang, Designing
  category-level attributes for discriminative visual recognition, in: IEEE
  Conference on Computer Vision and Pattern Recognition, 2013.

\bibitem{akata2013label}
Z.~Akata, F.~Perronnin, Z.~Harchaoui, C.~Schmid, Label-embedding for
  attribute-based classification, in: IEEE Conference on Computer Vision and
  Pattern Recognition, 2013.

\bibitem{sivic2003video}
J.~Sivic, A.~Zisserman, Video google: A text retrieval approach to object
  matching in videos, in: International Conference on Computer Vision, 2003.

\bibitem{jegou2010aggregating}
H.~J{\'e}gou, M.~Douze, C.~Schmid, P.~P{\'e}rez, Aggregating local descriptors
  into a compact image representation, in: IEEE Conference on Computer Vision
  and Pattern Recognition, 2010.

\bibitem{perronnin2010large}
F.~Perronnin, Y.~Liu, J.~S{\'a}nchez, H.~Poirier, Large-scale image retrieval
  with compressed fisher vectors, in: IEEE Conference on Computer Vision and
  Pattern Recognition, 2010.

\bibitem{ferrari2008learning}
V.~Ferrari, A.~Zisserman, Learning visual attributes, in: Conference on Neural
  Information Processing Systems, 2008.

\bibitem{siddiquie2011image}
B.~Siddiquie, R.~S. Feris, L.~S. Davis, Image ranking and retrieval based on
  multi-attribute queries, in: IEEE Conference on Computer Vision and Pattern
  Recognition, 2011.

\bibitem{kovashka2012whittlesearch}
A.~Kovashka, D.~Parikh, K.~Grauman, Whittlesearch: Image search with relative
  attribute feedback, in: IEEE Conference on Computer Vision and Pattern
  Recognition, 2012.

\bibitem{yu2012weak}
F.~X. Yu, R.~Ji, M.-H. Tsai, G.~Ye, S.-F. Chang, Weak attributes for
  large-scale image retrieval, in: IEEE Conference on Computer Vision and
  Pattern Recognition, 2012.

\bibitem{rastegari2012attribute}
M.~Rastegari, A.~Farhadi, D.~Forsyth, Attribute discovery via predictable
  discriminative binary codes, in: European Conference on Computer Vision,
  2012.

\bibitem{zhang2013semantic}
S.~Zhang, M.~Yang, X.~Wang, Y.~Lin, Q.~Tian, Semantic-aware co-indexing for
  image retrieval, in: International Conference on Computer Vision, 2013.

\bibitem{douze2011combining}
M.~Douze, A.~Ramisa, C.~Schmid, Combining attributes and fisher vectors for
  efficient image retrieval, in: IEEE Conference on Computer Vision and Pattern
  Recognition, 2011.

\bibitem{gordoa2012leveraging}
A.~Gordoa, J.~A. Rodr{\'\i}guez-Serrano, F.~Perronnin, E.~Valveny, Leveraging
  category-level labels for instance-level image retrieval, in: IEEE Conference
  on Computer Vision and Pattern Recognition, 2012.

\bibitem{naphade2006large}
M.~Naphade, J.~R. Smith, J.~Tesic, S.-F. Chang, W.~Hsu, L.~Kennedy,
  A.~Hauptmann, J.~Curtis, Large-scale concept ontology for multimedia,
  MultiMedia, IEEE 13~(3) (2006) 86--91.

\bibitem{LSVRC2010}
Large scale visual recognition challenge, \url{http://www.imagenet.
  org/challenges/LSVRC/2010} (2010).

\bibitem{bedagkar2014survey}
A.~Bedagkar-Gala, S.~K. Shah, A survey of approaches and trends in person
  re-identification, Image and Vision Computing 32~(4) (2014) 270--286.

\bibitem{vezzani2013people}
R.~Vezzani, D.~Baltieri, R.~Cucchiara, People reidentification in surveillance
  and forensics: A survey, ACM Computing Surveys (CSUR) 46~(2) (2013) 29.

\bibitem{gray2008viewpoint}
D.~Gray, H.~Tao, Viewpoint invariant pedestrian recognition with an ensemble of
  localized features, in: European Conference on Computer Vision, 2008.

\bibitem{ma2012local}
B.~Ma, Y.~Su, F.~Jurie, Local descriptors encoded by fisher vectors for person
  re-identification, in: European Conference on Computer Vision workshops,
  2012.

\bibitem{yang2014salient}
Y.~Yang, J.~Yang, J.~Yan, S.~Liao, D.~Yi, S.~Z. Li, Salient color names for
  person re-identification, in: European Conference on Computer Vision, 2014.

\bibitem{hirzer2012relaxed}
M.~Hirzer, P.~M. Roth, M.~K{\"o}stinger, H.~Bischof, Relaxed pairwise learned
  metric for person re-identification, in: European Conference on Computer
  Vision, 2012.

\bibitem{chen2015similarity}
D.~Chen, Z.~Yuan, G.~Hua, N.~Zheng, J.~Wang, Similarity learning on an explicit
  polynomial kernel feature map for person re-identification, in: IEEE
  Conference on Computer Vision and Pattern Recognition, 2015.

\bibitem{paisitkriangkrai2015learning}
S.~Paisitkriangkrai, C.~Shen, A.~v.~d. Hengel, Learning to rank in person
  re-identification with metric ensembles, in: IEEE Conference on Computer
  Vision and Pattern Recognition, 2015.

\bibitem{bazzani2013symmetry}
L.~Bazzani, M.~Cristani, V.~Murino, Symmetry-driven accumulation of local
  features for human characterization and re-identification, Computer Vision
  and Image Understanding 117~(2) (2013) 130--144.

\bibitem{zhao2013unsupervised}
R.~Zhao, W.~Ouyang, X.~Wang, Unsupervised salience learning for person
  re-identification, in: IEEE Conference on Computer Vision and Pattern
  Recognition, 2013.

\bibitem{zhao2014learning}
R.~Zhao, W.~Ouyang, X.~Wang, Learning mid-level filters for person
  re-identification, in: IEEE Conference on Computer Vision and Pattern
  Recognition, 2014.

\bibitem{liao2015person}
S.~Liao, Y.~Hu, X.~Zhu, S.~Z. Li, Person re-identification by local maximal
  occurrence representation and metric learning, in: IEEE Conference on
  Computer Vision and Pattern Recognition, 2015.

\bibitem{ahmed2015improved}
E.~Ahmed, M.~Jones, T.~K. Marks, An improved deep learning architecture for
  person re-identification, in: IEEE Conference on Computer Vision and Pattern
  Recognition, 2015.

\bibitem{shi2015transferring}
Z.~Shi, T.~M. Hospedales, T.~Xiang, Transferring a semantic representation for
  person re-identification and search, in: IEEE Conference on Computer Vision
  and Pattern Recognition, 2015.

\bibitem{zheng2015query}
L.~Zheng, S.~Wang, L.~Tian, F.~He, Z.~Liu, Q.~Tian, Query-adaptive late fusion
  for image search and person re-identification, in: IEEE Conference on
  Computer Vision and Pattern Recognition, 2015.

\bibitem{gray2007evaluating}
D.~Gray, S.~Brennan, H.~Tao, Evaluating appearance models for recognition,
  reacquisition, and tracking, in: IEEE International Workshop on Performance
  Evaluation for Tracking and Surveillance, 2007.

\bibitem{tao2015attributes}
R.~Tao, A.~W.~M. Smeulders, S.-F. Chang, Attributes and categories for generic
  instance search from one example, in: IEEE Conference on Computer Vision and
  Pattern Recognition, 2015.

\bibitem{sanchez2013image}
J.~S{\'a}nchez, F.~Perronnin, T.~Mensink, J.~Verbeek, Image classification with
  the fisher vector: theory and practice, International Journal of Computer
  Vision 105~(3) (2013) 222--245.

\bibitem{jegou2012aggregating}
H.~J{\'e}gou, F.~Perronnin, M.~Douze, J.~S{\'a}nchez, P.~P{\'e}rez, C.~Schmid,
  Aggregating local image descriptors into compact codes, IEEE Transactions on
  Pattern Analysis and Machine Intelligence 34~(9) (2012) 1704--1716.

\bibitem{krizhevsky2012imagenet}
A.~Krizhevsky, I.~Sutskever, G.~Hinton, Imagenet classification with deep
  convolutional neural networks, in: Conference on Neural Information
  Processing Systems, 2012.

\bibitem{razavian2014cnn}
A.~S. Razavian, H.~Azizpour, J.~Sullivan, S.~Carlsson, {CNN} features
  off-the-shelf: an astounding baseline for recognition, in: IEEE Conference on
  Computer Vision and Pattern Recognition Workshops, 2014.

\bibitem{babenko2014neural}
A.~Babenko, A.~Slesarev, A.~Chigorin, V.~Lempitsky, Neural codes for image
  retrieval, in: European Conference on Computer Vision, 2014.

\bibitem{girshick2014rich}
R.~Girshick, J.~Donahue, T.~Darrell, J.~Malik, Rich feature hierarchies for
  accurate object detection and semantic segmentation, in: IEEE Conference on
  Computer Vision and Pattern Recognition, 2014.

\bibitem{ng2015exploiting}
J.~Y.-H. Ng, F.~Yang, L.~S. Davis, Exploiting local features from deep networks
  for image retrieval, in: IEEE Conference on Computer Vision and Pattern
  Recognition Workshops, 2015.

\bibitem{berg2010automatic}
T.~L. Berg, A.~C. Berg, J.~Shih, Automatic attribute discovery and
  characterization from noisy web data, in: European Conference on Computer
  Vision, 2010.

\bibitem{shen2012mobile}
X.~Shen, Z.~Lin, J.~Brandt, Y.~Wu, Mobile product image search by automatic
  query object extraction, in: European Conference on Computer Vision, 2012.

\bibitem{perdoch2009efficient}
M.~Perdoch, O.~Chum, J.~Matas, Efficient representation of local geometry for
  large scale object retrieval, in: IEEE Conference on Computer Vision and
  Pattern Recognition, 2009.

\bibitem{van2010evaluating}
K.~van~de Sande, T.~Gevers, C.~G.~M. Snoek, Evaluating color descriptors for
  object and scene recognition, IEEE Transactions on Pattern Analysis and
  Machine Intelligence 32~(9) (2010) 1582--1596.

\bibitem{simonyan2014very}
K.~Simonyan, A.~Zisserman, Very deep convolutional networks for large-scale
  image recognition, in: International Conference on Learning Representations,
  2015.

\bibitem{perronnin2010improving}
F.~Perronnin, J.~S{\'a}nchez, T.~Mensink, Improving the fisher kernel for
  large-scale image classification, in: European Conference on Computer Vision,
  2010.

\bibitem{sharmanska2012augmented}
V.~Sharmanska, N.~Quadrianto, C.~H. Lampert, Augmented attribute
  representations, in: European Conference on Computer Vision, 2012.

\bibitem{von2007tutorial}
U.~Von~Luxburg, A tutorial on spectral clustering, Statistics and computing
  17~(4) (2007) 395--416.

\bibitem{huang2014circle}
J.~Huang, S.~Liu, J.~Xing, T.~Mei, S.~Yan, Circle \& search: Attribute-aware
  shoe retrieval, ACM Transactions on Multimedia Computing, Communications, and
  Applications (TOMM) 11~(1) (2014) 3.

\bibitem{zeiler2014visualizing}
M.~D. Zeiler, R.~Fergus, Visualizing and understanding convolutional networks,
  in: European Conference on Computer Vision, 2014.

\bibitem{pascal-voc-2007}
M.~Everingham, L.~Van~Gool, C.~K.~I. Williams, J.~Winn, A.~Zisserman, The
  {PASCAL} {V}isual {O}bject {C}lasses {C}hallenge 2007 {(VOC2007)} {R}esults,
  \url{http://www.pascal-network.org/challenges/VOC/voc2007/workshop/index.html}.

\bibitem{uijlings2013selective}
J.~R.~R. Uijlings, K.~van~de Sande, T.~Gevers, A.~W.~M. Smeulders, Selective
  search for object recognition, International Journal of Computer Vision
  104~(2) (2013) 154--171.

\bibitem{jegou2012negative}
H.~J{\'e}gou, O.~Chum, Negative evidences and co-occurences in image retrieval:
  The benefit of pca and whitening, in: European Conference on Computer Vision,
  2012.

\end{thebibliography}

\end{document}